\documentclass{article}

\usepackage{times}
\usepackage{graphicx} 

\usepackage{subfigure} 
\usepackage{bm}
\usepackage{natbib}
\usepackage{algorithm}
\usepackage{algorithmic}

\usepackage{hyperref}

\usepackage[accepted]{icml2015}

\usepackage{url}
\usepackage{amsmath}
\usepackage{amssymb}
\usepackage{cleveref}
\usepackage{macros}

\usepackage{pdfpages}
\icmltitlerunning{Predictive Entropy Search for Bayesian Optimization with Unknown Constraints}

\begin{document} 

\twocolumn[
\icmltitle{Predictive Entropy Search for Bayesian \\ 
            Optimization with Unknown Constraints}

\icmlauthor{Jos\'{e} Miguel Hern\'{a}ndez-Lobato$^1$}{jmh@seas.harvard.edu}
\icmladdress{Harvard University,
            Cambridge, MA 02138 USA}
\icmlauthor{Michael A. Gelbart$^1$}{mgelbart@seas.harvard.edu}
\icmladdress{Harvard University,
            Cambridge, MA 02138 USA}
\icmlauthor{Matthew W. Hoffman}{mwh30@cam.ac.uk}
\icmladdress{University of Cambridge,
            Cambridge, CB2 1PZ, UK}
\icmlauthor{Ryan P. Adams}{rpa@seas.harvard.edu}
\icmladdress{Harvard University,
            Cambridge, MA 02138 USA}
\icmlauthor{Zoubin Ghahramani}{zoubin@eng.cam.ac.uk}
\icmladdress{University of Cambridge,
            Cambridge, CB2 1PZ, UK}

\icmlkeywords{Bayesian optimization, entropy search, predictive entropy search, unknown constraints}

\vskip 0.3in
]

\begin{abstract}
Unknown constraints arise in many types of expensive black-box optimization problems. Several methods have been proposed recently for performing Bayesian optimization with constraints,  based on the expected improvement (EI) heuristic.  However, EI can lead to pathologies when used with constraints. For example, in the case of decoupled constraints---i.e., when one can independently evaluate the objective or the constraints---EI can encounter a pathology that prevents exploration. Additionally, computing EI requires a current best solution, which may not exist if none of the data collected so far satisfy the constraints. By contrast, information-based approaches do not suffer from these failure modes. In this paper, we present a new information-based method called Predictive Entropy Search with Constraints (PESC). We analyze the performance of PESC and show that it compares favorably to EI-based approaches on synthetic and benchmark problems, as well as several real-world examples. We demonstrate that PESC is an effective algorithm that provides a promising direction towards a unified solution for constrained Bayesian optimization.
\end{abstract} 

\footnotetext[1]{Authors contributed equally.}

\allowdisplaybreaks

\section{Introduction}
\vspace{-0.10cm}

We are interested in finding the global minimum $\xopt$ of an objective function $f(\x)$ over some bounded domain, typically ${\domain\subset\mathbb{R}^{d}}$, subject to the non-negativity of a series of constraint functions ${c_{1},\ldots,c_{K}}$. This can be formalized as
\begin{align}
\min_{\mathbf x\in\domain} f(\mathbf{x}) \quad\text{s.t.} \quad
c_{1}(\mathbf{x}) \geq 0,\ldots,c_{K}(\mathbf{x})\geq 0\,.\label{eq:problem}
\end{align}
However, $f$ and ${c_{1},\ldots,c_{K}}$
are unknown and can only be evaluated pointwise via expensive queries to black-boxes that
provide noise-corrupted evaluations of $f$ and $c_1, \ldots, c_K$.
We assume that~$f$ and each of the constraints~$c_k$ are defined over the entire space~$\domain$. We seek to find a solution to
(\ref{eq:problem}) with as few queries as possible. \emph{Bayesian optimization} \citep{Mockus1978} methods 
approach this type of problem by building a Bayesian model of the unknown objective
function and/or constraints, using this model to compute an \emph{acquisition function} that
represents how useful each input $\x$ is thought to be as a next evaluation, and then
maximizing this acquisition function to select a \emph{suggestion} for function evaluation. 

In this we work we extend Predictive Entropy Search (PES) \cite{hernandez2014} to solve (\ref{eq:problem}), an approach that we call Predictive Entropy Search with Constraints (PESC).  PESC is an acquisition function that approximates the expected information gain about the value of the constrained minimizer~$\xopt$. As we will show below, PESC is effective in practice and can be applied to a much wider variety of constrained problems than existing methods. 

\section{Related Work and Challenges}
\label{section:related-work}

Most previous approaches to Bayesian optimization with unknown
constraints are variants of expected improvement (EI) \cite{Mockus1978,Jones1998}. 
EI measures the expected amount by which observing at $\x$ leads to improvement over the current best value or \emph{incumbent} $\eta$:
\begin{equation}
\EI(\x|\eta,\mathcal{D})= \!\int\! \max(0, f(\mathbf{x}) - \eta) p(f(\mathbf{x})|\mathcal{D}) \,df(\mathbf{x}) \, ,
\end{equation}
where $\mathcal{D}$ is the collected data.

\subsection{Expected improvement with constraints} \label{sec:cei}

One way to use EI with constraints works by discounting EI by the posterior probability of a constraint violation. The resulting acquisition function, which we call expected improvement with constraints
(EIC), is given by
\begin{align}
    \alpha_\text{EIC}(\x)
    &=
    \EI(\x|\eta,\mathcal{D}^f)\prod_{k=1}^K p(c_k(\x)\geq 0|\mathcal{D}^k),
    \label{eq:EICacquisition}
\end{align}
where $\mathcal{D}^f$ is the set of objective function observations and $\mathcal{D}^k$ is the set of observations for constraint $k$.
Initially proposed by \citet{schonlau1998global}, EIC has recently been independently developed in \citet{snoek-2013a, Gelbart2014, Gardner2014}.
In the constrained case, $\eta$ is the smallest value of the posterior mean of $f$ such that all the constraints are satisfied at the corresponding location. 


\subsection{Augmented Lagrangian}\label{sec:lagrangian}

\citet{gramacy-augmented-lagrangian} propose a combination of 
the expected improvement heuristic and the augmented Lagrangian (AL)
optimization framework for constrained blackbox optimization.
AL methods are a class of algorithms for constrained nonlinear optimization
that work by iteratively optimizing the unconstrained AL:
\vspace{-0.5cm}

{\small
\begin{align}
L_A(\mathbf{x}|\lambda,p) = f(x) + \sum_{k=1}^K \left[ \frac{1}{2p} \min(0, c_k(\mathbf{x}))^2 - \lambda_k c_k(\mathbf{x}) \right]\nonumber
\end{align}
}where ${p > 0}$ is a penalty parameter and ${\lambda \geq 0}$ is an approximate Lagrange multiplier, both of which are updated at each iteration.

The method proposed by \citet{gramacy-augmented-lagrangian} uses Bayesian optimization with EI to solve the unconstrained \emph{inner} loop of the augmented Lagrangian formulation. 
AL is limited by requiring noiseless constraints so that $p$ and $\lambda$ can be updated at each iteration. 
In section \ref{sec:toy_problem} we show that PESC and EIC perform better than AL on the synthetic benchmark problem considered in \citet{gramacy-augmented-lagrangian}, even when the AL method has access to the true objective function and PESC and EIC do not.

\subsection{Integrated expected conditional improvement}

\citet{gramacy2010} propose an acquisition function based on the integrated
expected conditional improvement (IECI), which is given by
\begin{align}
\alpha_\text{IECI}(\x)=\int \left[ \EI(\x') - \EI(\x'|\x) \right] h(\mathbf{x}')d\mathbf{x}' \; ,
\label{eq:ieci}
\end{align}
where $\EI(\mathbf{x}')$ is the expected improvement at~$\mathbf{x}'$
and~$\EI(\mathbf{x}|\mathbf{x}')$ is the expected improvement at
$\mathbf{x}'$ when the objective has been evaluated at $\mathbf{x}$, but
without knowing the value obtained.  The IECI at $\mathbf{x}$ is the expected
reduction in improvement at $\mathbf{x}'$ under the density $h(\mathbf{x}')$ caused by
observing the objective at that location, where $h(\mathbf{x}')$ is the
probability of all the constraints being satisfied at $\mathbf{x}'$. 
\citet{Gelbart2014} compare IECI with EIC for optimizing the hyper-parameters of a
topic model with constraints on the entropy of the per-topic word distribution and
show that EIC outperforms IECI for this problem.

\subsection{Expected volume reduction}

\citet{picheny2014stepwise} proposes to sequentially explore the location that yields that largest the expected volume reduction (EVR) of the feasible region below the best feasible objective value $\eta$ found so far. This quantity is given by integrating the product of the probability of improvement and the probability of feasibility. 
That is,
\begin{align}
\alpha_\text{EVR}(\x) = - \int p[f(\x')\leq \min(\eta, f(\x))] h(\x')\del\x'\,,
\label{eq:ev}
\end{align}
where, as in IECI, $h(\x')$ is the probability that the constraints are satisfied at $\x'$. This step-wise uncertainty reduction approach is similar to PESC in that both methods work by
reducing a specific type of  uncertainty measure (entropy for PESC and expected volume for EVR). 

\subsection{Challenges}

EI-based methods for constrained optimization have several issues. First, when no point in the search space is feasible under the above definition, 
$\eta$ does not exist and the EI cannot be computed. This issue affects EIC, IECI, and EVR. To address this issue, \citet{Gelbart2014} modify EIC to ignore the factor~$\text{EI}(\x|\eta,\mathcal{D}^f)$ in (\ref{eq:EICacquisition})
and only consider the posterior probability of the constraints being satisfied when $\eta$ is not defined. The resulting acquisition function focuses only on searching for a feasible location and ignores learning about the objective $f$. 

Furthermore, \citet{Gelbart2014} identify a pathology with EIC when one is able to separately evaluate the objective or the constraints, i.e., the \emph{decoupled} case.  The best solution~$\x_\star$ must satisfy a conjunction of low objective value \emph{and} high (non-negative) constraint values.  By only evaluating the objective or a single constraint, this conjunction cannot be satisfied by a single observation under a myopic search policy. 
Thus, the new observed $\x$ cannot become the new incumbent as a result of a decoupled observation and the expected improvement is zero. Therefore standard EIC fails in the decoupled setting. \citet{Gelbart2014} circumvent this pathology
by treating decoupling as a special case and using a two-stage acquisition
function: first, $\x$ is chosen with EIC, and then, given $\x$, the task (whether to
evaluate the objective or one of the constraints) is chosen with the method in
\citet{VillemonteixVW09}. This approach does not take full
advantage of the available information in the way a joint selection of $\x$ and
the task would. Like EIC, the methods AL, IECI, and EVR are also not easily extended to the decoupled setting.


In addition to this difficulties, EVR and IECI are limited by having to compute the integrals in (\ref{eq:ieci}) and (\ref{eq:ev}) over the entire domain, which is done numerically over a grid on $\x'$ \cite{gramacy2010,picheny2014stepwise}.  The resulting acquisition function must then be globally optimized, which also requires
a grid on~$\x$. This nesting of grid operations limits the application of this method to small $\dimension$. 

Our new method, PESC, does not suffer from these pathologies.  First, the PESC acquisition function does not depend on the current best feasible solution, so it can operate coherently even when there is not yet a feasible solution. Second, PESC naturally separates the contribution of each task (objective or constraint) in its acquisition function. As a result, no pathology arises in the decoupled case and, thus, no \emph{ad hoc} modifications  to the acquisition function are required. Third, likewise EVR and IECI, PESC also involves computing a difficult integral (over the posterior on $\xopt$). However, this can be done efficiently using the sampling approach described in \citet{hernandez2014}.
Furthermore, in addition to its increased generality, our experiments show that PESC performs favorably when compared to EIC and AL even in the basic setting of joint evaluations to which these methods are most suited. 

\section{Predictive entropy search with constraints}
\label{sec:pesc}

We seek to maximize information about the location~$\mathbf{x}_{\star}$, the constrained global minimum, whose posterior distribution is~$p(\mathbf{x}_{\star}|\data^{0},\ldots,\data^{K})$. 
We assume that~$f$ and $c_{1},\ldots,c_{K}$ follow independent Gaussian process (GP) priors \citep[see, e.g.,][]{rasmussen2006gaussian} and that observation noise is
i.i.d. Gaussian with zero mean. GPs are widely-used probabilistic models for Bayesian nonparametric regression which provide a flexible framework for working with unknown response surfaces.

In the coupled setting we will let ${\data=\{(\mathbf{x}_n,\mathbf y_n)\}_{n\leq N}}$ denote all the observations up to step~$N$, where~$\mathbf y_n$ is a vector collecting the objective and constraint observations at step~$n$. The next query~$\mathbf{x}_{N+1}$ can then be defined as that which maximizes the expected reduction in the differential entropy~$\text{H}[\cdot]$ of the posterior on~$\mathbf{x}_\star$. 
We can write the PESC acquisition function as\begin{align}
\alpha(\mathbf{x}) & = 
\text{H}\left[\mathbf{x}_{\star}|\data\right]-
\mathbb{E}_{\mathbf y}\left\{\text{H}\left[\mathbf{x}_{\star}|\data\cup(\mathbf x, \mathbf y)\right]\right\}\label{eq:originalAcquisition}
\end{align}
where the expectation is taken with respect to
the posterior distribution on the noisy evaluations of~$f$ and~${c_1,\ldots,c_K}$ at~$\mathbf{x}$, that is,~$p(\mathbf y|\data, \mathbf x)$.

The exact computation of the above expression is infeasible in practice.
Instead, we follow \citet{houlsby2012,hernandez2014} and take advantage of the
symmetry of mutual information, rewriting this acquisition function as
the mutual information between $\mathbf y$ and
$\mathbf{x}_{\star}$ given the collected data $\data$. That is,
\begin{align}
\alpha(\mathbf{x}) & = 
 \text{H}\left[\mathbf y|\data, \mathbf x\right]-
\mathbb{E}_{\mathbf x_\star}\left\{\text{H}\left[
\mathbf y| \data, \mathbf x, \mathbf x_\star
\right]\right\}
\label{eq:newAcquisition}
\end{align}
where the expectation is now with respect to the posterior 
$p(\mathbf{x}_{\star}|\data)$
and where $p(\mathbf y|\data,\mathbf{x},\mathbf{x}_{\star})$
is the posterior predictive distribution for objective and constraint values given past data
and the location of the global solution to the constrained
optimization problem $\mathbf{x}_{\star}$. We call $p(\mathbf y|\data,\mathbf{x},\mathbf{x}_{\star})$ the \emph{conditioned predictive distribution} (CPD).

The first term on the right-hand side of (\ref{eq:newAcquisition}) is
straightforward to compute: it is the entropy of a product of
independent Gaussians, which is given by
\begin{equation}
\text{H}(\y|\mathcal{D},\mathbf{x}) = \log v_f+\sum_{k=1}^K \log v_k + \frac{K+1}{2}\log(2\pi e) \, ,
\label{eq:entropy-of-gaussians}
\end{equation}
where $v_f$ and $v_k$ are the predictive variances of the objective and constraints, respectively.
However, the second term in the right-hand side of (\ref{eq:newAcquisition}) has to
be approximated. For this, we first approximate the expectation by averaging
over samples of $\mathbf{x}_\star$ approximately drawn from
$p(\mathbf{x}_{\star}|\data)$. To sample $\mathbf{x}_\star$, we first
approximately draw $f$ and $c_1,\ldots,c_K$ from their GP posteriors using a
finite parameterization of these functions. Then we solve a constrained
optimization problem using the sampled functions to yield a sample of $\xopt$. This optimization approach is an extension
of the approach described in more detail by
\citet{hernandez2014}, extended to the constrained setting. For each value of
$\mathbf{x}_\star$ generated by this procedure, we approximate the CPD
$p(\mathbf y|\data,\mathbf{x},\mathbf{x}_{\star})$ as described in
the next section.

\subsection{Approximating the CPD}

Let ${\mathbf z = [f(\mathbf x), c_1(\mathbf x), \dots, c_K(\mathbf x)]^T}$
denote the concatenated vector of the noise-free objective and constraint
values at $\mathbf x$. We can approximate the CPD by first approximating the
posterior predictive distribution of $\mathbf z$ conditioned on $\data$,
$\mathbf x$, and $\mathbf{x}_{\star}$, which we call the \emph{noise free CPD}
(NFCPD), and then convolving that approximation with additive Gaussian noise of
variance $\sigma_0^2,\ldots,\sigma_K^2$. 

We first consider the distribution $p(\xopt\given\functions)$. The variable $\xopt$ is in fact a deterministic function of the latent functions $\functions$: in particular, $\xopt$ is the global minimizer if and only if (i) all constraints are satisfied at $\xopt$ and (ii) $f(\xopt)$ is the smallest feasible value in the domain. We can informally translate these deterministic conditions into a conditional probability: 
\vspace{-0.5cm}

{\small
\begin{equation}
\label{eq:pesc:x-star-given-functions}
p(\xopt\given\functions)= 
    \left[\prod_{k=1}^K \Theta\left[c_k(\xopt)\right]\right] \prod_{\x'\in\domain} \Psi(\x') \, ,
\end{equation}
}where $\Psi(\x')$ is defined as
\vspace{-0.5cm}

{\small
\begin{align*}
\left(\prod_{k=1}^K\Theta\left[c_k(\x')\right]\right)
    \Theta[f(\x')-f(\xopt)] + 
    \left(1 - \prod_{k=1}^K\Theta\left[c_k(\x')\right]\right)
\end{align*}
}and the symbol $\Theta$ denotes the Heaviside step function with the convention that ${\Theta(0)=1}$. 
The first product in (\ref{eq:pesc:x-star-given-functions}) encodes condition (i) and the infinite product over $\Psi(\x')$ encodes condition (ii).
Note that $\Psi(\x')$ also depends on $\xopt$ and $\functions$; we use the notation $\Psi(\x')$ for brevity.

Because $\z$ is simply a vector containing the values of $\functions$ at $\x$, $\z$ is also a deterministic function of $\functions$ and we can write $p(\z\given\functions,\x)$ using Dirac delta functions to pick out the values at $\x$:
\vspace{-0.5cm}

{\small
\begin{equation}
\label{eq:pesc:z-given-functions}
p(\z\given\functions,\x) = \delta[z_0-f(\x)] \prod_{k=1}^K \delta[z_k-c_k(\x)] \, .
\end{equation}
}We can now write the NFCPD by i) noting that $\mathbf{z}$ is independent of $\mathbf{x}^\star$ given $\functions$, ii)
multiplying the product of (\ref{eq:pesc:x-star-given-functions}) and (\ref{eq:pesc:z-given-functions}) by $p(f,c_1,\dots,c_K|\data)$
and iii) integrating out the latent functions $\functions$:
\vspace{-0.5cm}

{\small
\begin{multline}
p(\mathbf z|\data,\mathbf{x},\mathbf{x}_{\star}) \propto
\int \delta[z_0-f(\mathbf x)] \left[
    \textstyle\prod_{k=1}^K \delta[z_k-c_k(\mathbf x)]\right] \\
    \left[\textstyle\prod_{k=1}^K \Theta\left[c_k(\xopt)\right]\right] 
    \left[ \textstyle\prod_{\mathbf x'\neq \x}  \Psi(\x') \right] \Psi(\x) \\
p(f,c_1,\dots,c_K|\data) \,df\,dc_1\dots\,dc_k \,,
\label{eq:NFCPD}
\end{multline}
}where $p(f,c_1,\dots,c_K|\data)$ is an infinite-dimensional Gaussian given by the GP  posterior on $f,c_1,\ldots,c_K$,
and we have separated $\Psi(\x)$ out from the infinite product over $\x'$.

We find a Gaussian approximation to (\ref{eq:NFCPD}) in several steps. The general approach is to separately approximate the factors that do and do not depend on $\x$, so that the computations associated with the latter factors can be reused rather than recomputed for each $\x$. In (\ref{eq:NFCPD}), the only factors that depend on $\x$ are the deltas in the first line, and $\Psi(\x)$.

Let $\mathbf{f}$ denote the $(N+1)$-dimensional vector containing objective function evaluations at $\mathbf x_\star$ and $\mathbf{x}_1,\ldots,\mathbf{x}_N$, and define constraint vectors $\mathbf{c}_1,\ldots,\mathbf{c}_K$ similarly. 
Then, we approximate (\ref{eq:NFCPD}) by conditioning only on $\mathbf{f}$ and $\mathbf{c}_1,\ldots,\mathbf{c}_K$, rather than the full $\functions$.
We first approximate the factors in (\ref{eq:NFCPD}) that do not depend on $\x$ as
\vspace{-0.5cm}

{\small
\begin{multline}
    q_1(\mathbf f, \mathbf c_1, \dots, \mathbf c_K) = \\
    \left[\textstyle\prod_{k=1}^K \Theta[c_{k0}]\right] 
    \left[ \textstyle\prod_{n=1}^N \Psi(\x_n) \right]
    p(\mathbf{f},\mathbf c_1,\dots,\mathbf c_K|\data)
    \label{eq:NFCPDfinite}
\end{multline}
}where $p(\mathbf{f},\mathbf c_1,\dots,\mathbf c_K|\data)$ is the GP
predictive distribution for objective and constraint values. Because
(\ref{eq:NFCPDfinite}) is not tractable, we approximate the normalized version
of $q_1$ with a product of Gaussians using expectation propagation (EP)
\cite{minka2001family}. In particular, we obtain
\begin{multline}
\textstyle
Z^{-1}_{1} q_1(\mathbf{f},\mathbf{c}_1,\ldots,\mathbf{c}_K) 
\approx  q_2(\mathbf{f},\mathbf{c}_1,\ldots,\mathbf{c}_K) =  \\ 
\textstyle \mathcal{N}(\mathbf{f}|\mathbf{m}_0,\mathbf{V}_0)
\prod_{k=1}^K \mathcal{N}(\mathbf{c}_k|\mathbf{m}_k,\mathbf{V}_k)\,,
\label{eq:gaussian-q}
\end{multline}
where $Z_{1}$ is the normalization constant of $q_1$ and $(\mathbf m_k,
\mathbf V_k)$ for ${k=0,\dots,K}$ are the mean and covariance terms determined by
EP. See the supplementary material for details on the EP approximation. Roughly speaking, EP 
approximates each true (but intractable) factor in (\ref{eq:NFCPDfinite}) with a 
Gaussian factor whose parameters are iteratively refined. The product of all these
Gaussian factors produces a tractable Gaussian approximation to (\ref{eq:NFCPDfinite}).

We now approximate the portion of (\ref{eq:NFCPD}) that does depend on $\x$, namely the first line and the factor $\Psi(\x)$, by replacing the deltas with $p(\mathbf z|\mathbf f, \mathbf c_1, \dots, \mathbf c_K)$, the ${K+1}$ dimensional, Gaussian conditional
distribution given by the GP priors on~${f,c_1,\ldots,c_K}$. Our full approximation to (\ref{eq:NFCPD}) is then

\vspace{-0.5cm}
{\small
\begin{multline}
    p(\mathbf z|\data,\mathbf x, \mathbf x_\star)
    \approx Z_2^{-1} \int 
    \textstyle
    p(\mathbf z|\mathbf f, \mathbf c_1, \dots, \mathbf c_K) 
    \Psi(\x) \\
    q_2(\mathbf{f},\mathbf{c}_1,\ldots,\mathbf{c}_K)\,d\mathbf{f}\,d\mathbf{c}_1\,\cdots\,d\mathbf{c}_K\,,
    \label{eq:NFCPDapproximation}
\end{multline}
}where~$Z_2$ is a normalization constant. 
From here, we analytically marginalize out all integration variables except $f_0 = f(\xopt)$; see the supplementary material for the full details. This calculation, and those that follow, must be repeated for every $\x$; however, the EP approximation in (\ref{eq:gaussian-q}) can be reused over all $\x$. After performing the integration, we arrive at 

\vspace{-0.5cm}
{\small
\begin{align}
    &p(\mathbf z|\data,\mathbf x, \mathbf x_\star) \approx \nonumber\\ &
    \frac{1}{Z_3} \int \Psi(\x)
    \mathcal{N}([ z_0, f_0 ]|\mathbf{m}_0',\mathbf{V}_0') 
    \prod_{k=1}^K \mathcal{N}(z_k|m_k',v_k') \,df_0
    \,,\label{eq:NFCPDapproximationMarginalized}
\end{align}
}where $z_0=f(\mathbf{x})$. Details on how to compute the means $m_1',\ldots,m_K'$ and variances
$v_1',\ldots,v_K'$, as well as the 2-dimensional mean vector $\mathbf{m}_0'$
and the $2\times 2$ covariance matrix $\mathbf{V}_0'$ can be found in the
supplementary material.

We perform one final approximation to (\ref{eq:NFCPDapproximationMarginalized}).
We approximate this distribution with a product of independent Gaussians that have the same marginal means and variances as
(\ref{eq:NFCPDapproximationMarginalized}). This corresponds to a single iteration of EP; see the supplementary material for details.


\subsection{The PESC acquisition function}

By approximating the NFCPD with a product of independent Gaussians,
we can approximate the entropy in the CPD by performing the following operations.
First, we add the noise variances 
to the marginal variances of our final approximation of the NFCPD and second, we compute the entropy with (\ref{eq:entropy-of-gaussians}).
The PESC acquisition function, which approximates (\ref{eq:originalAcquisition}), is then
\vspace{-0.5cm}

{\small
\begin{multline}
\alpha_\text{PESC}(\x) = 
\left\{ \log v_f^\text{PD}(\x)+ \sum_{k=1}^K \log v_k^\text{PD}(\x)  \right\} - \\
\frac{1}{M}\sum_{m=1}^{M} \left\{ 
\log v_f^\text{CPD}\left(\x\given\xopt^{(m)}\right) + 
\sum_{k=1}^K \log v_k^\text{CPD}\left(\x\given\xopt^{(m)}\right) \right\}\,,
\label{eq:pesc_acquisition_function}
\end{multline}
}where $M$ is the number of samples drawn from
$p(\mathbf{x}_{\star}|\data)$, $\xopt^{(m)}$ is the $m$-th of these samples, $v_f^\text{PD}(\mathbf{x})$ and $v_k^\text{PD}(\mathbf{x})$ are
the predictive variances for the noisy evaluations of $f$ and $c_k$ at $\mathbf{x}$, respectively, and
$v_f^\text{CPD}(\mathbf{x}|\xopt^{(m)})$ and
$v_k^\text{CPD}(\mathbf{x}|\xopt^{(m)})$ are
the approximated marginal variances of the CPD for the noisy evaluations of~$f$ and~$c_k$ at $\mathbf{x}$
given that ${\mathbf{x}_\star = \xopt^{(m)}}$.
Marginalization of (\ref{eq:pesc_acquisition_function}) over the GP hyper-parameters 
can be done efficiently as in \citet{hernandez2014}.

The PESC acquisition function is additive in the expected amount of information that is obtained from
the evaluation of each task (objective or constraint) at any particular location $\mathbf{x}$.
For example, the expected information gain obtained from the evaluation of $f$ at $\mathbf{x}$
is given by the term~${\frac{1}{M}\sum_{m=1}^{M}\left[\log v_f^\text{PD}(\mathbf{x}) - \log v_f^\text{CPD}(\mathbf{x}|\mathbf{x}_\star^{(m)})\right]}$ in
(\ref{eq:pesc_acquisition_function}).
The other~$K$ terms in (\ref{eq:pesc_acquisition_function}) measure the
corresponding contribution from evaluating each of the constraints.
This allows PESC to easily address the decoupled scenario when one can 
independently evaluate the different functions at different locations.
In other words, Equation (\ref{eq:pesc_acquisition_function}) is a sum of individual
acquisition functions, one for each function that we can evaluate.
Existing methods for Bayesian optimization with unknown constraints (described in Section \ref{section:related-work}) do not possess this desirable property.
Finally, the complexity of PESC is of order $O(MKN^3)$ per iteration in the coupled setting.
As with unconstrained PES, this is dominated by the cost of a matrix inversion
in the EP step.

\begin{figure*}[t]
\centering

\subfigure[Marginal posteriors]{
  \includegraphics[height=0.24\textwidth]{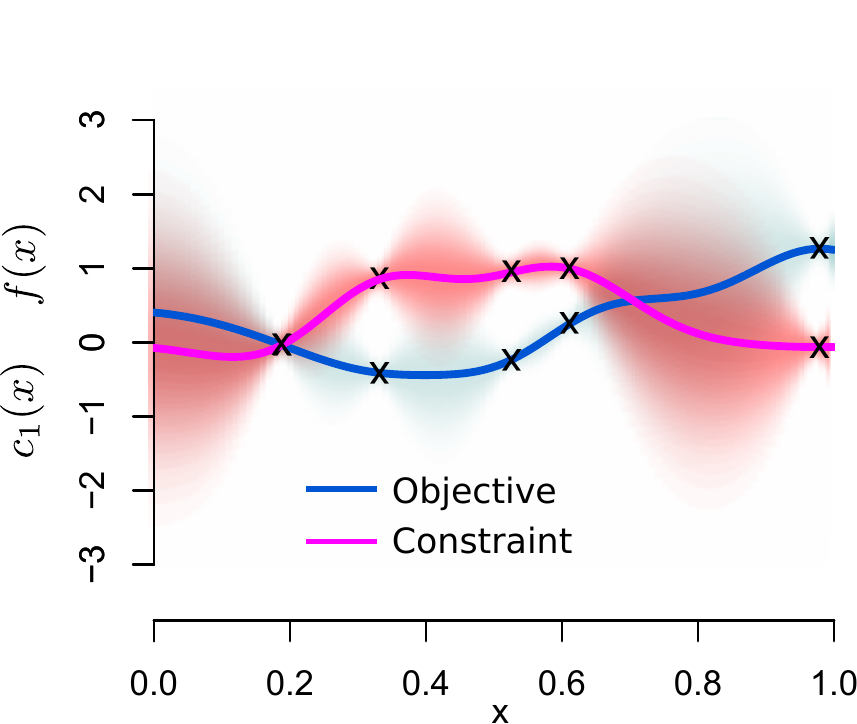}
  \label{fig:experiments:pesc:accuracy-marginal-posteriors}
}
\subfigure[Acquisition functions]{
  \includegraphics[height=0.24\textwidth]{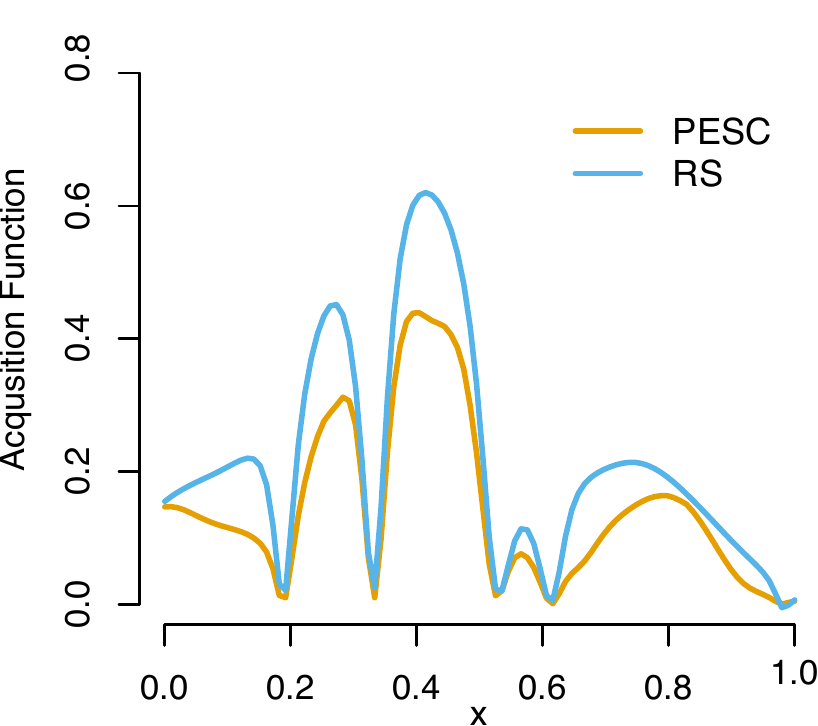}
  \label{fig:experiments:pesc:accuracy-acquisition-functions} 
}
\subfigure[Performance in $1\dimension$]{
  \includegraphics[height=0.24\textwidth]{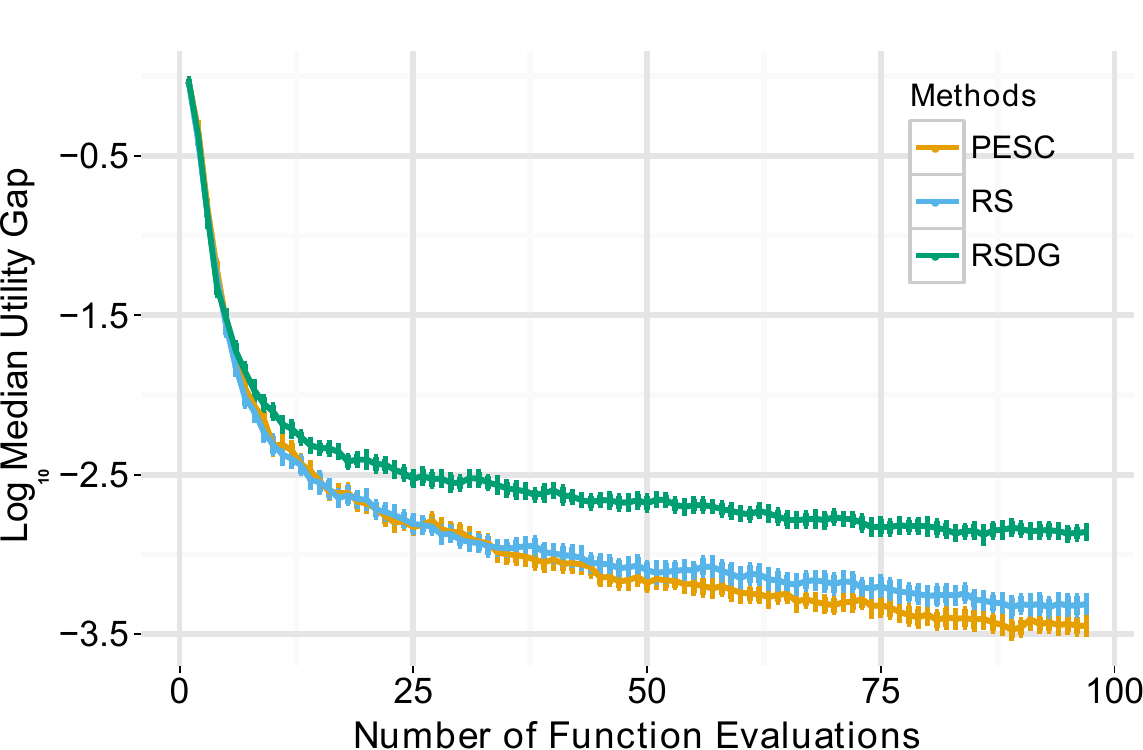}
  \label{fig:experiments:pesc:accuracy-performance} 
}

\caption[Assessing the accuracy of the PESC approximation.]{Assessing the accuracy of the PESC approximation. (a) Marginal posterior predictive distributions for the objective and constraint given some collected data denoted by $\times$'s. (b) PESC and RS acquisition functions given the data in (a). (c) Median utility gap for PESC, RS and RSDG in the experiments with synthetic functions sampled from the GP prior with $\dimension=1$.}
\label{fig:accuracy_approximation}
\end{figure*}

\section{Experiments}\label{sec:experiments}

We evaluate the performance of PESC through experiments with i) synthetic functions sampled from the GP prior distribution,
ii) analytic benchmark problems previously used in the literature on
Bayesian optimization with unknown constraints and iii) real-world constrained optimization problems. 

For case i) above, the synthetic functions sampled from the GP prior are
generated following the same experimental set up as in
\citet{hennig-schuler-2012} and \citet{hernandez2014}. The search space is the unit
hypercube of dimension $d$, and the ground truth objective $f$ is a sample from
a zero-mean GP with a squared exponential covariance function of unit amplitude
and length scale ${\ell = 0.1}$ in each dimension. We represent the function $f$
by first sampling from the GP prior on a grid of 1000 points generated using a
Halton sequence \citep[see][]{leobacher-2014} and then defining $f$ as the resulting GP posterior mean. We
use a single constraint function $c_1$ whose ground truth is sampled in the
same way as $f$.  The evaluations for $f$ and $c_1$ are contaminated with
i.i.d. Gaussian noise with variance ${\sigma_f^2 = \sigma_1^2 =0.01}$. 

\subsection{Accuracy of the PESC approximation}\label{sec:approximation_quality}

We first analyze the accuracy of the approximation to~(\ref{eq:newAcquisition}) generated by PESC.
We compare the PESC approximation with a ground truth for~(\ref{eq:newAcquisition})
obtained by rejection sampling (RS). The RS method works by discretizing the
search space using a uniform grid. The expectation with respect to~$p(\mathbf{x}_{\star}|\mathcal{D}_{n})$
in (\ref{eq:newAcquisition}) is then approximated by Monte Carlo. To achieve this, $f$ and $c_1,\ldots,c_K$ are sampled
on the grid and the grid cell with positive $c_1,\ldots,c_K$
(feasibility) and the lowest value of $f$ (optimality) is
selected.  For each sample of $\mathbf{x}_\star$ generated by this procedure,~$\text{H}\left[p(\mathbf y|\mathcal{D}_{n},\mathbf{x},\mathbf{x}_{\star})\right]$
is approximated by rejection sampling: we select those samples of $f$ and $c_1,\ldots,c_K$ 
whose corresponding feasible optimal solution is the sampled $\mathbf{x}_\star$ and reject the other samples.
We then assume that the selected samples for $f$ and $c_1,\ldots,c_K$ are independent and have
Gaussian marginal distributions. Under this assumption, 
$\text{H}\left[p(\mathbf y|\mathcal{D}_{n},\mathbf{x},\mathbf{x}_{\star})\right]$
can be approximated using the formula for the entropy of independent Gaussian random variables,
with the variance parameters in this formula being equal to the empirical marginal variances of the
selected samples of $f$ and $c_1,\ldots,c_K$ at $\mathbf{x}$ plus the corresponding noise variances
$\sigma_f^2$ and $\sigma_1^2,\ldots,\sigma_K^2$.

The left plot in Figure \ref{fig:accuracy_approximation} shows the posterior
distribution for $f$ and $c_1$ given 5 evaluations sampled from the GP prior with ${d=1}$.
The posterior is computed using the optimal GP hyperparameters.
The corresponding approximations to (\ref{eq:newAcquisition}) generated by PESC and RS
are shown in the middle plot of Figure \ref{fig:accuracy_approximation}.
Both PESC and RS use a total of 50 samples from
$p(\mathbf{x}_{\star}|\mathcal{D}_{n})$
when approximating the expectation in (\ref{eq:newAcquisition}).
The PESC approximation is very accurate, and importantly
its maximum value is very close to the maximum value of the RS approximation.

One disadvantage of the RS method is its high cost, which scales with the size of the grid used.
This grid has to be large to guarantee good performance, especially when $d$ is large.
An alternative is to use a small dynamic grid that changes as data is collected.
Such a grid can be obtained by sampling from
$p(\mathbf{x}_{\star}|\mathcal{D}_{n})$
using the same approach as in PESC. The samples obtained would then form the
dynamic grid. The resulting method is called Rejection Sampling with a Dynamic Grid (RSDG).

We compare the performance of PESC, RS and RSDG in experiments with synthetic
data corresponding to 500 pairs of $f$ and $c_1$ sampled from the GP prior with ${d=1}$.
At each iteration, RSDG draws the same number of samples of $\mathbf{x}_\star$ as PESC.
We assume that the GP hyperparameter values are known to each method.
Recommendations are made by finding the location with lowest
posterior mean for $f$ such that $c_1$ is non-negative with probability at least
${1-\delta_1}$, where ${\delta_1 = 0.05}$. For reporting purposes, we set the utility $u(\mathbf{x})$ of a
recommendation $\mathbf{x}$ to be $f(\mathbf{x})$ if~${\mathbf{x}}$
satisfies the constraint, and otherwise a penalty value of the worst (largest)
objective function value achievable in the search space. 
For each recommendation at $\mathbf{x}$, we compute
the utility gap $|u(\mathbf{x}) - u(\mathbf{x}_\star)|$, where $\mathbf{x}_\star$ is the true solution of the optimization problem. Each method is initialized with the same three random points drawn with Latin hypercube sampling.

\begin{figure*}[t]
\centering

\subfigure[Optimizing GP samples in $\dimension=2$]{
  \includegraphics[height=0.21\textwidth]{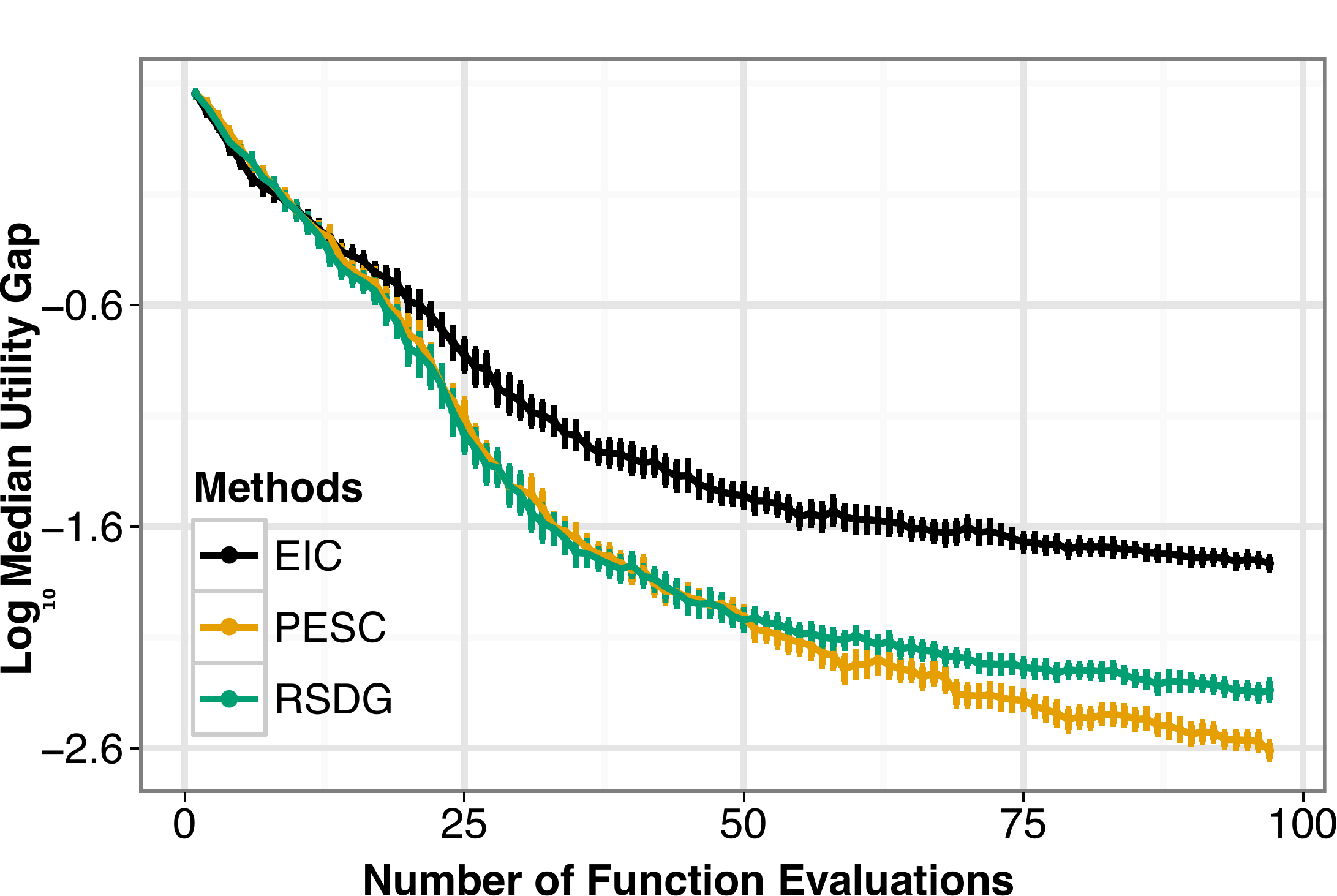}
}
\subfigure[Optimizing GP samples in $\dimension=8$]{
	\includegraphics[height=0.21\textwidth]{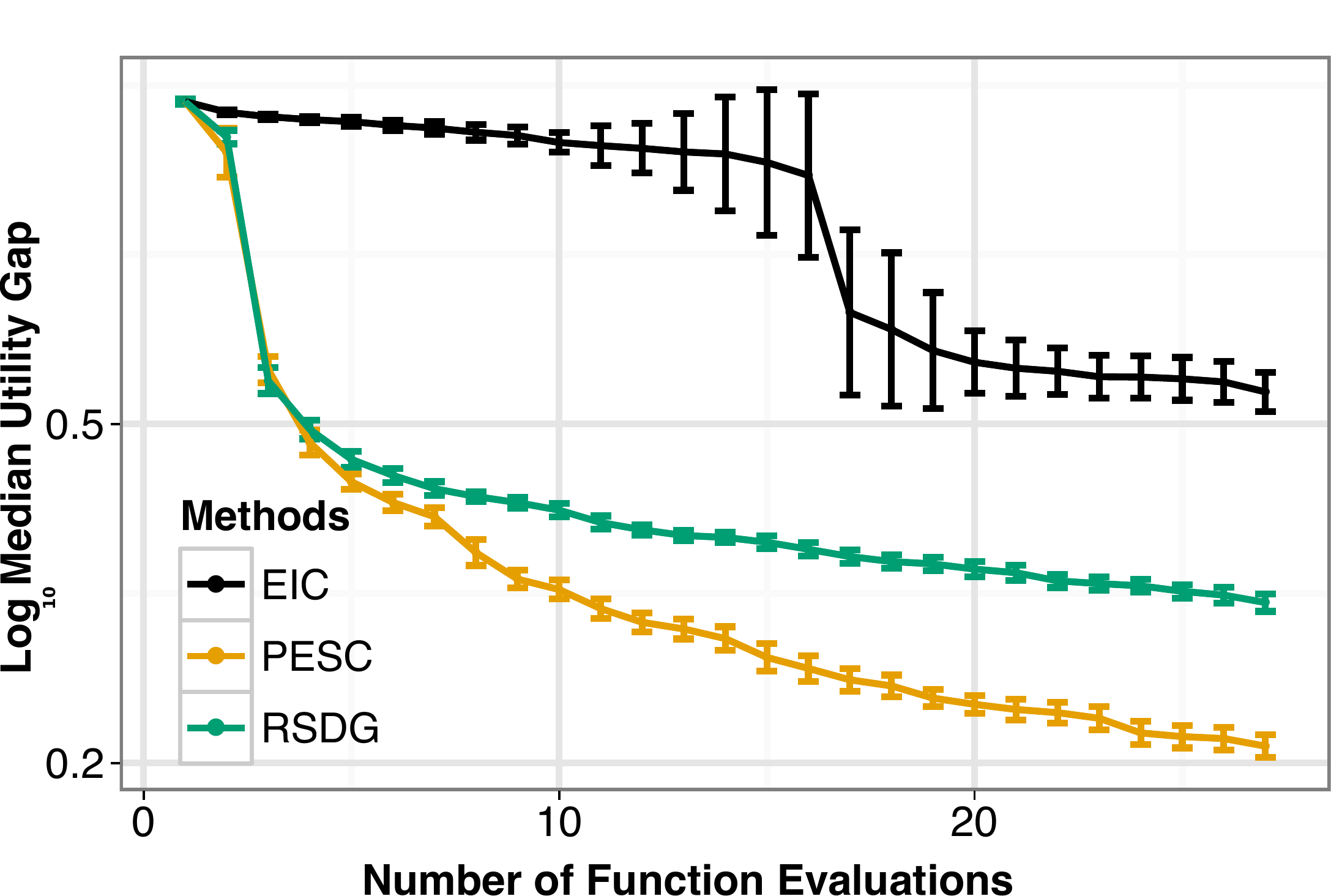}
}
\subfigure[2$\dimension$ toy problem]{
	\includegraphics[height=0.21\textwidth]{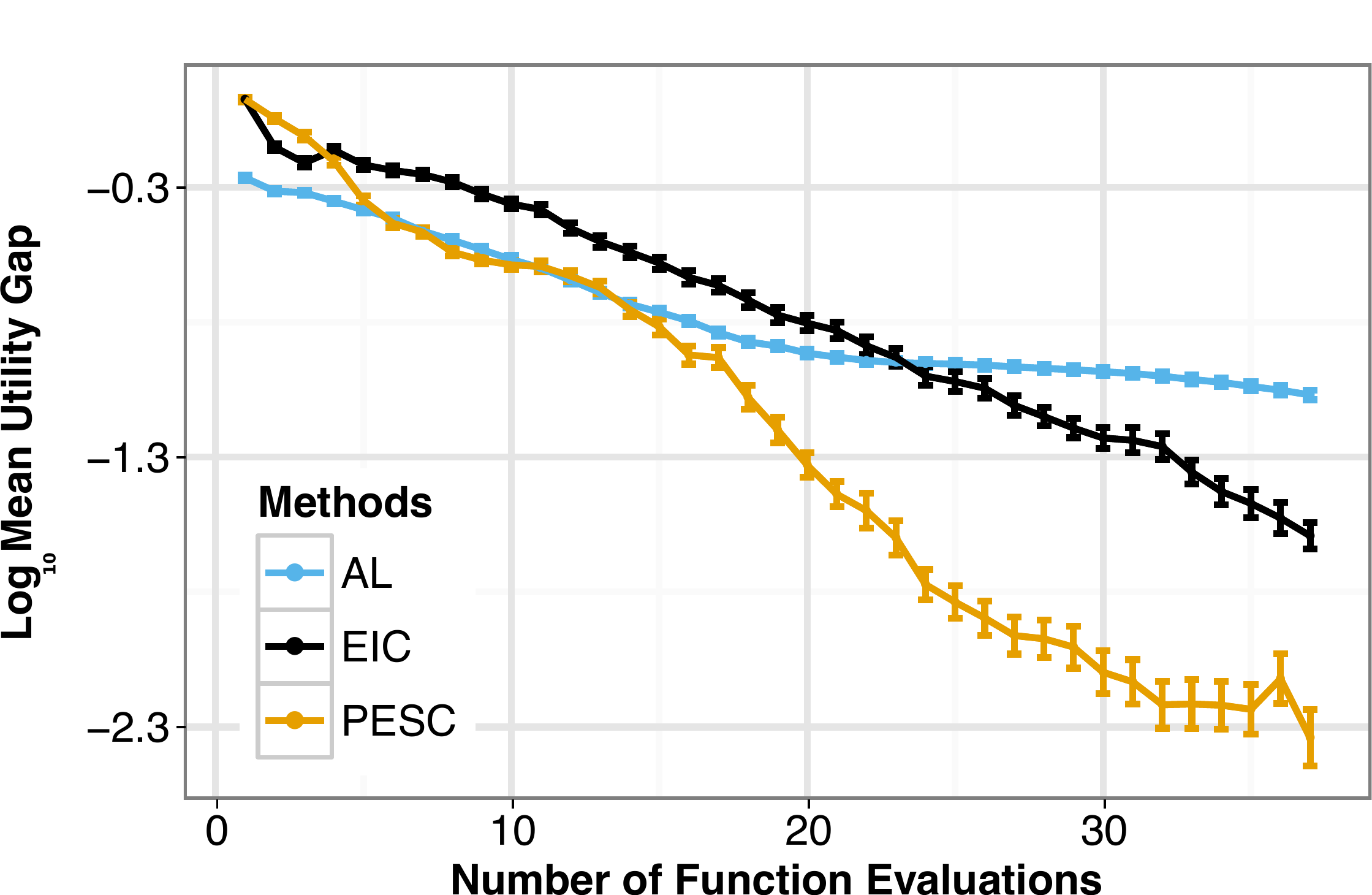}
	\label{fig:toy-example}
}
\label{fig:synthetic-and-toy}
\caption{Assessing PESC on synthetic problems. (a,b) Compare PESC to EIC and RSDG on optimizing samples from the GP in dimension 2 and 8 respectively, and (c) compares PESC to AL and EIC.}
\end{figure*}

The right plot in Figure \ref{fig:accuracy_approximation} shows the median of
the utility gap for each method across the 500 realizations of $f$ and~$c_1$.
The $x$-axis in this plot is the number of joint function evaluations for $f$
and $c_1$.  We report the median because the empirical distribution of the
utility gap is heavy-tailed and in this case the median is more representative of the
location of the bulk of the data than the mean. The heavy tails arise
because we are measuring performance across 500 different
optimization problems with very different degrees of difficulty. 
In this and all following experiments,
standard errors on the reported plot are computed using the bootstrap.
The plot shows that PESC and RS are better than RSDG. Furthermore, PESC is very
similar to RS, with PESC even performing slightly better at the end of the data collection
process since PESC is not limited by a finite grid as RS is. These results show that
PESC yields a very accurate approximation of the information gain. Furthermore, although RSDG performs worse than PESC,
RSDG is faster because the rejection sampling operation (with a small grid) 
is less expensive than the EP algorithm. Thus, RSDG is an attractive alternative to PESC when
the available computing time is very limited.



\subsection{Synthetic functions in 2 and 8 input dimensions}\label{sec:d_2_and_8}

We also compare the performance of PESC and RSDG with that of EIC (Section \ref{sec:cei}) using the
same experimental protocol as in the previous section, but with dimensionalities~${\dimension=2}$ and~${\dimension=8}$.  
We do not compare with RS here because its use of grids does not scale to higher dimensions.
Figure \ref{fig:synthetic-and-toy} shows the utility gap for each
method across 500 different samples of $f$ and $c_1$ from the GP prior with~${\dimension=2}$ (a) and~${\dimension=8}$ (b).
Overall, PESC is the best method, followed by RSDG and EIC.
RSDG performs similarly to PESC when~${\dimension=2}$, but is significantly worse when~${\dimension=8}$.
This shows that, when $\dimension$ is high, grid based approaches (e.g. RSDG) are
at a disadvantage with respect to methods that do not require a grid (e.g. PESC).

\subsection{A toy problem}\label{sec:toy_problem}

We compare PESC with EIC and AL (Section \ref{sec:lagrangian}) in the toy problem described in \citet{gramacy-augmented-lagrangian}.
We seek to minimize the function $f(x) = x_1 + x_2$, subject to the constraint functions $c_1(\x) \geq 0$ and $c_2(\x) \geq 0$, given by
\begin{align}
c_1(\x) &= 0.5\sin{(2\pi(x_1^2-2x_2))}+x_1+2x_2-1.5\,,\\
c_2(\x) &= -x_1^2-x_2^2 + 1.5\,,
\end{align}
where $\x$ is confined to the unit square. The evaluations for $f$, $c_1$ and
$c_2$ are noise-free. We compare PESC and EIC with $\delta_1 = \delta_2 = 0.025$ and
a squared exponential GP kernel.
PESC uses 10 samples from $p(\mathbf{x}_{\star}|\mathcal{D}_{n})$
when approximating the expectation in (\ref{eq:newAcquisition}).
We use the AL implementation
provided by \citet{gramacy-augmented-lagrangian} in the R package \emph{laGP} which is based on the squared exponential kernel and assumes the objective $f$ is known. Thus, in order for this implementation to be used, AL has an advantage over other methods in that it has access to the true objective function.
 In all three methods, the GP hyperparameters are estimated by maximum likelihood.

Figure \ref{fig:toy-example} shows the mean utility gap for each method
across 500 independent realizations. Each realization
corresponds to a different initialization of the methods with three data points selected with Latin hypercube sampling.  Here, we report the
mean because we are now measuring performance across
realizations of the same optimization problem and the heavy-tailed effect
described in Section \ref{sec:approximation_quality} is less severe. The
results show that PESC is significantly better than EIC and AL for this
problem.  EIC is superior to AL, which performs slightly better 
at the beginning, presumably because it has access to the ground truth objective $f$.

\subsection{Finding a fast neural network}\label{sec:nnet}
In this experiment, we tune the hyperparamters of a three-hidden-layer neural network subject
to the constraint that the prediction time must not exceed 2 ms on a GeForce GTX 580 GPU (also used for training).
The search space consists of 12 parameters: 2 learning rate parameters (initial and decay rate), 
2 momentum parameters (initial and final), 2 dropout parameters (input layer and other layers),
2 other regularization parameters (weight decay and max weight norm), the number of hidden units
in each of the 3 hidden layers, the activation function (RELU or sigmoid). The network is trained
using the \emph{deepnet} package\footnote{https://github.com/nitishsrivastava/deepnet}, and the prediction time is computed as the average time of 1000 predictions, each for a batch of size 128. The network is trained on the MNIST digit classification task with momentum-based stochastic gradient descent for 5000 iterations. The objective is reported as the classification error rate on the validation set. As above, we treat constraint violations as the worst possible value (in this case a classification error of 1.0).

\begin{figure*}[t]
\centering
\subfigure[Tuning a fast neural network]{
\includegraphics[width=0.85\columnwidth]{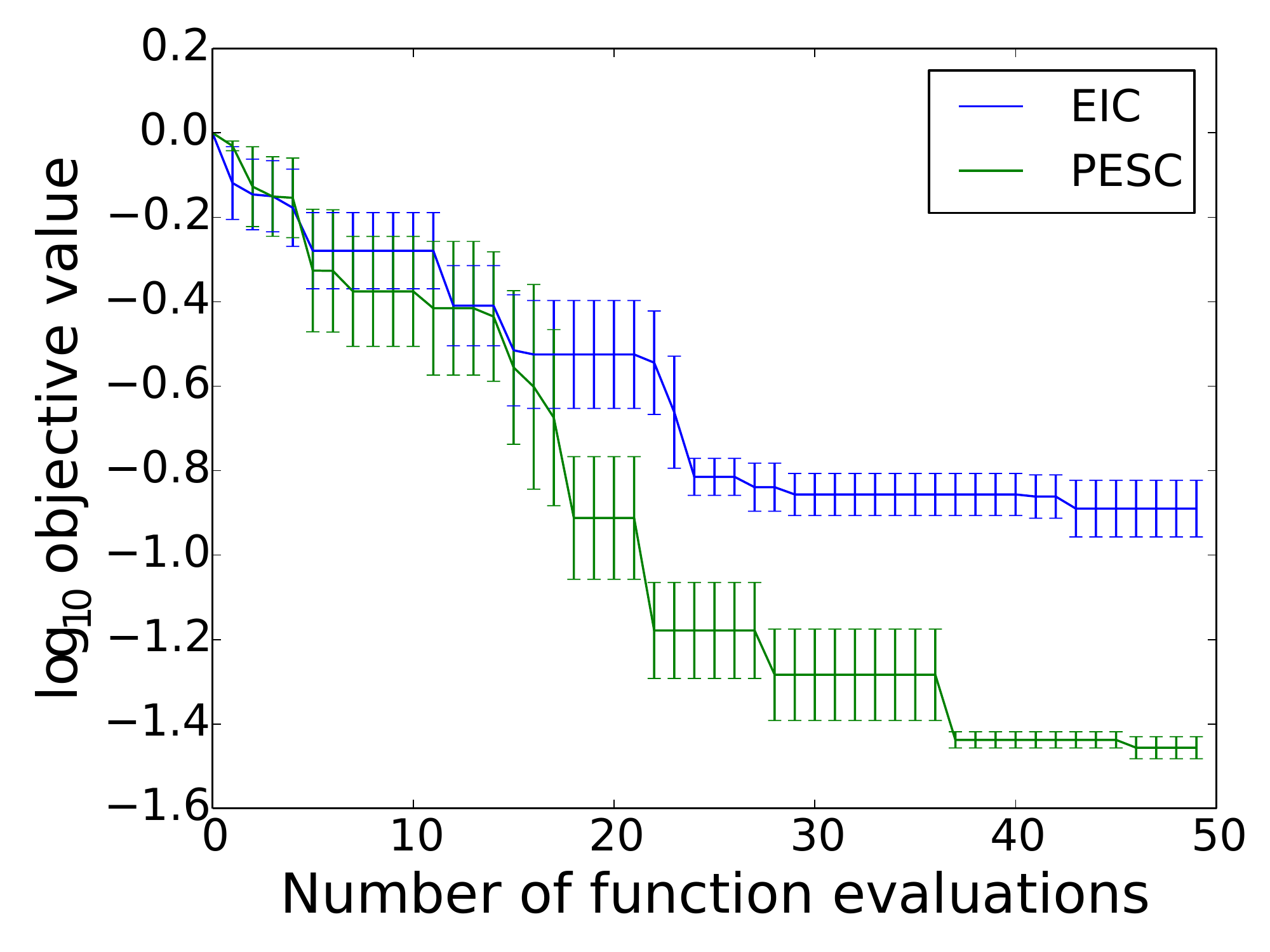}
\label{fig:net-coupled}
}
\subfigure[Tuning Hamiltonian MCMC]{
\includegraphics[width=0.86\columnwidth]{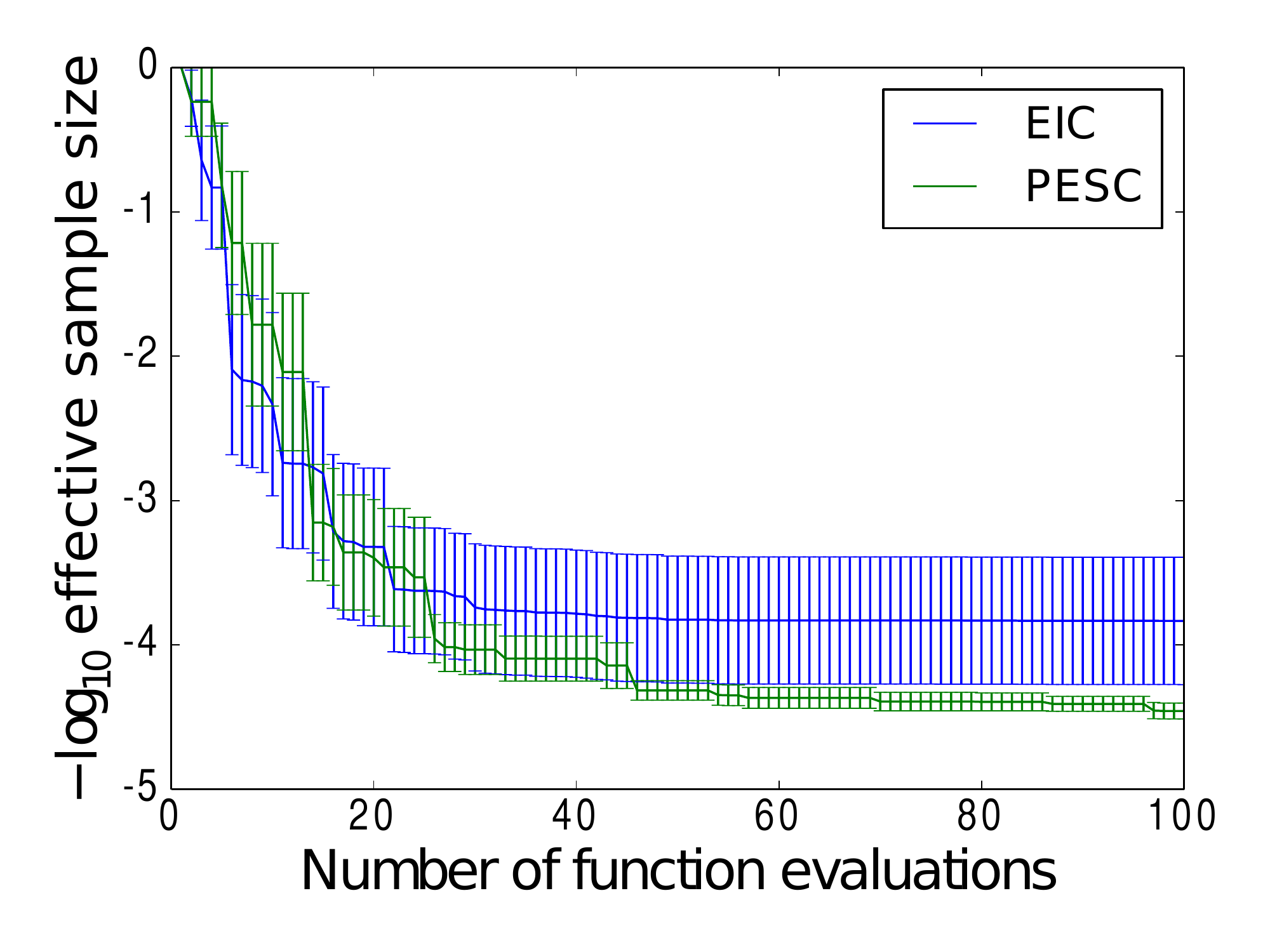}
\label{fig:hmc}
}
\caption{Comparing PESC and EIC for (a) minimizing classification error of a 3-hidden-layer neural network constrained to make predictions in under 2 ms, and (b) tuning Hamiltonian Monte Carlo to maximize the number of effective samples within 5 minutes of compute time.}
\end{figure*}

Figure~\ref{fig:net-coupled} shows the results of 50 iterations of Bayesian optimization. In this experiment and the next, the $y$-axis represents observed objective values, ${\delta_1=0.05}$, a Mat\'{e}rn 5/2 GP covariance kernel is used, and GP hyperparameters are integrated out using slice sampling \citep{Neal00slicesampling} as in \citet{snoek-etal-2012b}. Curves are the mean over 5 independent experiments. We find that PESC performs significantly better than EIC. However, when the noise level is high, reporting the best objective observation is an overly optimistic metric (due to ``lucky'' evaluations); on the other hand, ground-truth is not available. Therefore, to validate our results further, we used the recommendations made at the final iteration of Bayesian optimization for each method (EIC and PESC) and evaluted the function with these recommended parameters. We repeated the evaluation 10 times for each of the 5 repeated experiments to compute a ground-truth score averaged of 50 function evaluations. This procedure yields a score of $7.0 \pm 0.6 \%$ for PESC and $49 \pm 4 \%$ for EIC (as in the figure, constraint violations are treated as a classification error of $100\%$). This result is consistent with Figure~\ref{fig:net-coupled} in that PESC performs significantly better than EIC, but also demonstrates that, due to noise, Figure~\ref{fig:net-coupled} is overly optimistic. While we may believe this optimism to affect both methods equally, the ground-truth measurement provides a more reliable result and a much clearer understanding of the classification error attained by Bayesian optimization.

\subsection{Tuning Markov chain Monte Carlo}\label{sec:hmc}
Hybrid Monte Carlo, also known as Hamiltonian Monte Carlo (HMC), is a popular Markov Chain Monte Carlo (MCMC) technique that uses gradient information in a numerical integration to select the next sample. However, using numerical integration gives rise to new parameters like the integration step size and the number of integration steps. Following the experimental set up in \citet{Gelbart2014}, we optimize the number of effective samples produced by an HMC sampler limited to 5 minutes of computation time, subject to passing of the Geweke \citep{Geweke92} and Gelman-Rubin \citep{gelmanrubin} convergence diagnostics, as well as the constraint that the numerical integration should not diverge. We tune 4 parameters of an HMC sampler: the integration step size, number of integration steps, fraction of the allotted 5 minutes spent in burn-in, and an HMC mass parameter \citep[see][]{nealbook}. We use the \emph{coda} R package \citep{rcoda} to compute the effective sample size and the Geweke convergence diagnostic, and the \emph{PyMC} python package \citep{PyMC} to compute the Gelman-Rubin diagnostic over two independent traces. Following \citet{Gelbart2014}, we impose the constraints that the absolute value of the Geweke test score be at most 2.0 and the Gelman-Rubin score be at most 1.2, and sample from the posterior distribution of a logistic regression problem using the UCI German credit data set \citep{Frank2010}. 

Figure~\ref{fig:hmc} evaluates EIC and PESC on this task, averaged over 10 independent experiments. 
As above, we perform a ground-truth assessment of the final recommendations. The average effective sample size is $3300 \pm 1200$ for PESC and $2300 \pm 900$ for EIC. From these results we draw a similar conclusion to that of Figure~\ref{fig:hmc}; namely, that PESC outperforms EIC but only by a small margin, and furthermore that the experiment is very noisy.

\section{Discussion}


In this paper, we addressed global optimization with unknown constraints. Motivated by the weaknesses of existing methods, we presented PESC, a method based on the theoretically appealing expected information gain heuristic. We showed 
that the approximations in PESC are quite accurate, and that PESC performs about equally well to a ground truth method based on rejection sampling.  In \cref{sec:d_2_and_8,sec:toy_problem,sec:nnet,sec:hmc}, we showed that PESC outperforms current methods such as EIC and AL over a variety of problems. Furthermore, PESC is easily applied to problems with decoupled constraints, without additional computational cost or the pathologies discussed in \citet{Gelbart2014}.

One disadvantage of PESC is that it is relatively difficult to implement: in
particular, the EP approximation often leads to numerical instabilities.
Therefore, we have integrated our implementation, which carefully addresses
these numerical issues, into the open-source Bayesian optimization package \emph{Spearmint} at
\url{https://github.com/HIPS/Spearmint/tree/PESC}. We have demonstrated that PESC is a
flexible and powerful method and we hope the existence of such a method will
bring constrained Bayesian optimization into the standard toolbox of Bayesian
optimization practitioners.


\subsection*{Acknowledgements}

Jos\'e Miguel Hern\'andez-Lobato acknowledges support from the Rafael del Pino Foundation.
Zoubin Ghahramani acknowledges support from Google Focused Research Award and EPSRC grant EP/I036575/1.
Matthew W. Hoffman acknowledges support from EPSRC grant EP/J012300/1.



\section*{Supplementary material}

# Predictive Entropy Search for Bayesian Optimization with Unknown Constraints
## *Supplementary Material*

José Miguel Hernández-Lobato,* Harvard University, USA
Michael A. Gelbart*, Harvard University, USA
Matthew W. Hoffman, University of Cambridge, UK
Ryan P. Adams, Harvard University, USA
Zoubin Ghahramani, University of Cambridge, UK

May 12, 2015

## 1   Description of Expectation Propagation

PESC computes a Gaussian approximation to the NFCPD (main text, Eq. (11)) using *Expectation Propagation* (EP) (Minka, 2001). EP is a method for approximating a product of factors (often a single prior factor and multiple likelihood factors) with a tractable distribution, for example a Gaussian. EP generates a Gaussian approximation by approximating each individual factor with a Gaussian. The product all these Gaussians results in a single Gaussian distribution that approximates the product of all the exact factors. This is in contrast to the Laplace approximation which fits a single Gaussian distribution to the whole posterior. EP can be intuitively understood as fitting the individual Gaussian approximations by minimizing the Kullback-Leibler (KL) divergences between each exact factor and its corresponding Gaussian approximation. This would correspond to matching first and second moments between exact and approximate factors. However, EP does this moment matching in the *context* of all the other approximate factors, since we are ultimately interested in having a good approximation in regions where the overall posterior probability is high. Concretely, assume we wish to approximate the distribution

$$q(\mathbf{x}) = \prod_{n=1}^{N} q_n(\mathbf{x})$$

with the approximate distribution

$$\tilde{q}(\mathbf{x}) = \prod_{n=1}^{N} \tilde{q}_n(\mathbf{x}) \,, \tag{1}$$

where each $\tilde{q}_n(\mathbf{x})$ is Gaussian with specific parameters. Consider now that we wish to tune the parameters of a particular approximate factor $\tilde{q}_n(\mathbf{x})$. Then, we define the *cavity* distribution $\tilde{q}^{\backslash n}(\mathbf{x})$ as

$$\tilde{q}^{\backslash n}(\mathbf{x}) = \prod_{n' \neq n}^{N} \tilde{q}_{n'}(\mathbf{x}) = \frac{\tilde{q}(\mathbf{x})}{\tilde{q}_n(\mathbf{x})} \,. \tag{2}$$

Instead of matching the moments of $q_n(\mathbf{x})$ and $\tilde{q}_n(\mathbf{x})$, EP matches the moments (minimizes the KL divergence) of $q_n(\mathbf{x})\tilde{q}^{\backslash n}(\mathbf{x})$ and $\tilde{q}_n(\mathbf{x})\tilde{q}^{\backslash n}(\mathbf{x}) = \tilde{q}(\mathbf{x})$. This causes the approximation quality to be higher in places where the

---

*Equal authors

entire distribution $\tilde{q}(\mathbf{x})$ is high, at the expense of approximation quality in less relevant regions where $\tilde{q}(\mathbf{x})$ is close to zero. To compute the moments of $q_n(\mathbf{x})\hat{q}^{\setminus n}(\mathbf{x})$ we use Eqs. (5.12) and (5.13) in Minka (2001), which give the first and second moments of a distribution $p(\mathbf{x})\mathcal{N}(\mathbf{x}\,|\,\mathbf{m},\mathbf{V})$ in terms of the derivatives of its log partition function.

Thus, given some initial value for the parameters of all the $\tilde{q}_n(\mathbf{x})$, the steps performed by EP are

1. Choose an $n$.

2. Compute the cavity distribution $\hat{q}^{\setminus n}(\mathbf{x})$ given by Eq. (2) using the formula for dividing Gaussians.

3. Compute the first and second moments of $q_n(\mathbf{x})\hat{q}^{\setminus n}(\mathbf{x})$ using Eqs. (5.12) and (5.13) in Minka (2001). This yields an updated Gaussian approximation $\tilde{q}(\mathbf{x})$ to $q(\mathbf{x})$ with mean and variance given by these moments.

4. Update $\tilde{q}_n(\mathbf{x})$ as the ratio between $\tilde{q}(\mathbf{x})$ and $\hat{q}^{\setminus n}(\mathbf{x})$, using the formula for dividing Gaussians.

5. Repeat steps 1 to 4 until convergence.

The EP updates for all the $q_n(\mathbf{x})$ maybe be done in parallel by performing steps 1 to 4 for $n = 1,\ldots,N$ using the same $\tilde{q}(\mathbf{x})$ for each $n$. After his, the new $\tilde{q}(\mathbf{x})$ is computed, according to Eq. (1), as the product of all the newly updated $q_n(\mathbf{x})$. For these latter computations, one uses the formula for multiplying Gaussians.

## 2 The Gaussian approximation to the NFCPD

In this section we fill in the details of computing a Gaussian approximation to $q(\mathbf{f},\mathbf{c}_1,\ldots,\mathbf{c}_K)$, given by Eq. (12) of the main text. Recall that $\mathbf{f} = (f(\mathbf{x}_\star),f(\mathbf{x}_1),\ldots,f(\mathbf{x}_n))$ and $\mathbf{c}_k = (c(\mathbf{x}_\star),c(\mathbf{x}_1),\ldots,c(\mathbf{x}_n))$, where $k = 1,\ldots,K$, and the first element in these vectors is accessed with the index number 0 and the last one with the index number $n$. The expression for $q(\mathbf{f},\mathbf{c}_1,\ldots,\mathbf{c}_K)$ can be written as

$$q(\mathbf{f},\mathbf{c}_1,\ldots,\mathbf{c}_K) = \frac{1}{Z_q}\mathcal{N}\left(\mathbf{f}\,|\,\mathbf{m}_{\text{pred}}^{\mathbf{f}},\mathbf{V}_{\text{pred}}^{\mathbf{f}}\right)\left[\prod_{k=1}^{K}\mathcal{N}\left(\mathbf{c}_k\,|\,\mathbf{m}_{\text{pred}}^{\mathbf{c}_k},\mathbf{V}_{\text{pred}}^{\mathbf{c}_k}\right)\right]\times$$
$$\left[\prod_{k=1}^{K}\Theta[c_{k,0}]\right]\prod_{n=1}^{N}\left[\left\{\prod_{k=1}^{K}\Theta[c_{k,n}]\right\}\Theta[f_n-f_0]+\left\{1-\prod_{k=1}^{K}\Theta[c_{k,n}]\right\}\right],\tag{3}$$

where $\mathbf{m}_{\text{pred}}^{f}$ and $\mathbf{V}_{\text{pred}}^{f}$ are the mean and covariance matrix of the posterior distribution of $\mathbf{f}$ given the data in $\mathcal{D}^f$ and $\mathbf{m}_{\text{pred}}^{\mathbf{c}_k}$ and $\mathbf{V}_{\text{pred}}^{\mathbf{c}_k}$ are the mean and covariance matrix of the posterior distribution of $\mathbf{c}_k$ given the data in $\mathcal{D}^k$. In particular, from Rasmussen & Williams (2006) Eqs. 2.22-2.24, we have that

$$\begin{aligned}\mathbf{m}_{\text{pred}}^{\mathbf{f}} &= \mathbf{K}_\star^f(\mathbf{K}^f+\nu_f^2\mathbb{I})^{-1}\mathbf{y}^f\,,\\\mathbf{V}_{\text{pred}}^{\mathbf{f}} &= \mathbf{K}_{\star,\star}^f-\mathbf{K}_\star^f(\mathbf{K}^f+\nu_f^2\mathbb{I})^{-1}[\mathbf{K}_\star^f]^\top\,,\end{aligned}$$

where $\mathbf{K}_\star^f$ is an $(N+1)\times N$ matrix with the prior cross-covariances between elements of $\mathbf{f}$ and $f_1,\ldots,f_n$ and $\mathbf{K}_{\star,\star}^f$ is an $(N+1)\times(N+1)$ matrix with the prior covariances between elements of $\mathbf{f}$ and $\nu_f$ is the standard deviation of the additive Gaussian noise in the evaluations of $f$. Similarly, we have that

$$\begin{aligned}\mathbf{m}_{\text{pred}}^{\mathbf{c}_k} &= \mathbf{K}_\star^k(\mathbf{K}^k+\nu_k^2\mathbb{I})^{-1}\mathbf{y}^k\,,\\\mathbf{V}_{\text{pred}}^{\mathbf{c}_k} &= \mathbf{K}_{\star,\star}^k-\mathbf{K}_\star^k(\mathbf{K}^k+\nu_k^2\mathbb{I})^{-1}[\mathbf{K}_\star^k]^\top\,,\end{aligned}$$

where $\mathbf{K}_\star^k$ is an $(N+1)\times N$ matrix with the prior cross-covariances between elements of $\mathbf{c}_k$ and $c_{k,1},\ldots,c_{k,n}$ and $\mathbf{K}_{\star,\star}^k$ is an $(N+1)\times(N+1)$ matrix containing the prior covariances between the elements of $\mathbf{c}_k$ and $\nu_k$ is the standard deviation of the additive Gaussian noise in the evaluations of $c_k$. We will refer to the non-Gaussian factors in Eq. (3) as

$$h_n(f_n,f_0,c_{1,n},\ldots,c_{k,n}) = \left\{\prod_{k=1}^{K}\Theta\left[c_{k,n}\right]\right\}\Theta\left[f_n-f_0\right]+\left\{1-\prod_{k=1}^{K}\Theta\left[c_{k,n}\right]\right\}\tag{4}$$

(this is called $\Psi(\mathbf{x}_n, \mathbf{x}_\star, \mathbf{f}, \mathbf{c}_1, \ldots, \mathbf{c}_K)$ in the main text) and

$$g_k(c_{k,0}) = \Theta[c_{k,0}],\tag{5}$$

such that Eq. (3) can be written as

$$q(\mathbf{f}, \mathbf{c}_1, \ldots, \mathbf{c}_K) \propto \mathcal{N}(\mathbf{f} \mid \mathbf{m}_{\text{pred}}^{\mathbf{f}}, \mathbf{V}_{\text{pred}}^{\mathbf{f}}) \left[ \prod_{k=1}^{K} \mathcal{N}(\mathbf{c}_k \mid \mathbf{m}_{\text{pred}}^{\mathbf{c}_k}, \mathbf{V}_{\text{pred}}^{\mathbf{c}_k}) \right]$$

$$\left[ \prod_{n=1}^{N} h_n(f_n, f_0, c_{1,n}, \ldots, c_{k,n}) \right] \left[ \prod_{k=1}^{K} g_k(c_{k,0}) \right].\tag{6}$$

We then approximate the exact non-Gaussian factors $h_n(f_n, f_0, c_{1,n}, \ldots, c_{k,n})$ and $g_k(c_{k,0})$ with corresponding Gaussian approximate factors $\tilde{h}_n(f_n, f_0, c_{1,n}, \ldots, c_{k,n})$ and $\tilde{g}_k(c_{k,0})$, respectively. By the assumed independence of the objective and constraints, these approximate factors take the form

$$\tilde{h}_n(f_n, f_0, c_{1,n}, \ldots, c_{k,n}) \propto \exp\left\{ -\frac{1}{2}[f_n \, f_0] \mathbf{A}_{h_n} [f_n \, f_0]^\top + [f_n \, f_0] \mathbf{b}_{h_n} \right\}$$

$$\prod_{k=1}^{K} \exp\left\{ -\frac{1}{2} a_{h_n} c_{k,n}^2 + b_{h_n} c_{k,n} \right\}\tag{7}$$

and

$$\tilde{g}_k(c_{k,0}) \propto \exp\left\{ -\frac{1}{2} a_{g_k} c_{k,0}^2 + b_{g_k} c_{k,0} \right\},\tag{8}$$

where $\mathbf{A}_{h_n}$, $\mathbf{b}_{h_n}$, $a_{h_n}$, $b_{h_n}$, $a_{g_k}$ and $b_{g_k}$ are the natural parameters of the respective Gaussian distributions.

The right-hand side of Eq. (6) is then approximated by

$$\tilde{q}(\mathbf{f}, \mathbf{c}_1, \ldots, \mathbf{c}_K) \propto \mathcal{N}\left(\mathbf{f} \mid \mathbf{m}_{\text{pred}}^{f}, \mathbf{V}_{\text{pred}}^{f}\right) \left[ \prod_{k=1}^{K} \mathcal{N}\left(\mathbf{c}_k \mid \mathbf{m}_{\text{pred}}^{\mathbf{c}_k}, \mathbf{V}_{\text{pred}}^{\mathbf{c}_k}\right) \right]$$

$$\left[ \prod_{n=1}^{N} \tilde{h}_n(f_n, f_0, c_{1,n}, \ldots, c_{k,n}) \right] \left[ \prod_{k=1}^{K} \tilde{g}_k(c_{k,0}) \right].\tag{9}$$

Because the $\tilde{h}_n(f_n, f_0, c_{1,n}, \ldots, c_{k,n})$ and $\tilde{g}_k(c_{k,0})$ are Gaussian, they can be combined with the Gaussian terms in the first line of Eq. (9) and written as

$$\tilde{q}(\mathbf{f}, \mathbf{c}_1, \ldots, \mathbf{c}_K) \propto \mathcal{N}(\mathbf{f} \mid \mathbf{m}^{\mathbf{f}}, \mathbf{V}^{\mathbf{f}}) \left[ \prod_{k=1}^{K} \mathcal{N}(\mathbf{c}_k \mid \mathbf{m}^{\mathbf{c}_k}, \mathbf{V}^{\mathbf{c}_k}) \right],\tag{10}$$

where, by applying the formula for products of Gaussians, we obtain

$$\mathbf{V}^{\mathbf{f}} = \left[ \left(\mathbf{V}_{\text{pred}}^{\mathbf{f}}\right)^{-1} + \widetilde{\mathbf{V}}^{\mathbf{f}} \right]^{-1}, \qquad\qquad \mathbf{m}^{\mathbf{f}} = \mathbf{V}^{\mathbf{f}} \left[ \left(\mathbf{V}_{\text{pred}}^{\mathbf{f}}\right)^{-1} \mathbf{m}_{\text{pred}}^{\mathbf{f}} + \widetilde{\mathbf{m}}^{\mathbf{f}} \right],$$

$$\mathbf{V}^{\mathbf{c}_k} = \left[ \left(\mathbf{V}_{\text{pred}}^{\mathbf{c}_k}\right)^{-1} + \widetilde{\mathbf{V}}^{\mathbf{c}_k} \right]^{-1}, \qquad\qquad \mathbf{m}^{\mathbf{c}_k} = \mathbf{V}^{\mathbf{c}_k} \left[ \left(\mathbf{V}_{\text{pred}}^{\mathbf{c}_k}\right)^{-1} \mathbf{m}_{\text{pred}}^{\mathbf{c}_k} + \widetilde{\mathbf{m}}^{\mathbf{c}_k} \right],\tag{11}$$

with the following definitions for $\widetilde{\mathbf{V}}$, $\widetilde{\mathbf{m}}^{\mathbf{f}}$, $\widetilde{\mathbf{V}}^{\mathbf{c}_k}$, and $\widetilde{\mathbf{m}}^{\mathbf{c}_k}$:

- $\widetilde{\mathbf{V}}^{\mathbf{f}}$ is an $(N+1) \times (N+1)$ precision matrix with entries given by

  - $\tilde{v}_{n,n}^{\mathbf{f}} = [\mathbf{A}_{h_n}]_{1,1}$ for $n = 1, \ldots, N$,

  - $\tilde{v}_{0,n}^{\mathbf{f}} = \tilde{v}_{n,0}^{\mathbf{f}} = [\mathbf{A}_{h_n}]_{1,2}$ for $n = 1, \ldots, N$,

- $\tilde{v}_{0,0}^{\mathbf{f}} = \sum_{n=1}^{N} [\mathbf{A}_{h_n}]_{2,2}$,
    - and all other entries are zero.

- $\tilde{\mathbf{m}}^{\mathbf{f}}$ is an $(N+1)$-dimensional vector with entries given by
    - $\tilde{m}_n^{\mathbf{f}} = [\mathbf{b}_{h_n}]_1$ for $n = 1, \ldots, N$,
    - $\tilde{m}_0^{\mathbf{f}} = \sum_{n=1}^{N} [\mathbf{b}_{h_n}]_2$.

- $\tilde{\mathbf{V}}^{\mathbf{c}_k}$ is a $(N+1) \times (N+1)$ diagonal precision matrix with entries given by
    - $\tilde{v}_{n,n}^{\mathbf{c}_k} = a_{h_n}$ for $n = 1, \ldots, N$,
    - $\tilde{v}_{0,0}^{\mathbf{c}_k} = a_{g_k}$.

- $\tilde{\mathbf{m}}^{\mathbf{c}_k}$ is an $(N+1)$-dimensional vector such that
    - $\tilde{m}_n^{\mathbf{c}_k} = b_{h_n}$ for $n = 1, \ldots, N$,
    - $\tilde{m}_0^{\mathbf{c}_k} = b_{g_k}$.

In the next section we explain how to actually obtain the values of $\mathbf{A}_{h_n}, \mathbf{b}_{h_n}, a_{h_n}, b_{h_n}, a_{g_k}$ and $b_{g_k}$ by running EP.

# 3   The EP approximation to $h_n$ and $g_k$

In this section we explain how to iteratively refine the approximate Gaussian factors $\tilde{h}_n$ and $\tilde{g}_k$ with EP.

## 3.1   EP update operations for $\tilde{h}_n$

EP adjusts each $\tilde{h}_n$ by minimizing the KL divergence between

$$h_n(f_n, f_0, c_{1,n}, \ldots, c_{k,n}) \tilde{q}^{\backslash n}(\mathbf{f}, \mathbf{c}_1, \ldots, \mathbf{c}_K) \tag{12}$$

and

$$\tilde{h}_n(f_n, f_0, c_{1,n}, \ldots, c_{k,n}) \tilde{q}^{\backslash n}(\mathbf{f}, \mathbf{c}_1, \ldots, \mathbf{c}_K), \tag{13}$$

where we define the cavity distribution $\tilde{q}^{\backslash n}(\mathbf{f}, \mathbf{c}_1, \ldots, \mathbf{c}_K)$ as

$$\tilde{q}^{\backslash n}(\mathbf{f}, \mathbf{c}_1, \ldots, \mathbf{c}_K) = \frac{\tilde{q}(\mathbf{f}, \mathbf{c}_1, \ldots, \mathbf{c}_K)}{\tilde{h}_n(f_n, f_0, c_{1,n}, \ldots, c_{k,n})}.$$

Because $\tilde{q}(\mathbf{f}, \mathbf{c}_1, \ldots, \mathbf{c}_K)$ and $\tilde{h}_n(f_n, f_0, c_{1,n}, \ldots, c_{k,n})$ are both Gaussian, $\tilde{q}^{\backslash n}(\mathbf{f}, \mathbf{c}_1, \ldots, \mathbf{c}_K)$ is also Gaussian. If we marginalize out all variables except those which $\tilde{h}_n$ depends on, namely $f_n, f_0, c_{1,n}, \ldots, c_{k,n}$, then we have that $\tilde{q}^{\backslash n}(f_n, f_0, c_{1,n}, \ldots, c_{k,n})$ takes the form

$$\tilde{q}^{\backslash n}(f_n, f_0, c_{1,n}, \ldots, c_{k,n}) \propto \mathcal{N}\left([f_n f_0] \,|\, \mathbf{m}_{n,\text{old}}^{f_n, f_0}, \mathbf{V}_{n,\text{old}}^{f_n, f_0}\right) \left[\prod_{k=1}^{K} \mathcal{N}\left(c_{k,n} \,|\, m_{n,\text{old}}^{c_{k,n}}, v_{n,\text{old}}^{c_{k,n}}\right)\right], \tag{14}$$

where, by applying the formula for dividing Gaussians, we obtain

$$\mathbf{V}_{n,\text{old}}^{f_n, f_0} = \left\{\left[\mathbf{V}_{f_n, f_0}^{\mathbf{f}}\right]^{-1} - \mathbf{A}_{h_n}\right\}^{-1}, \tag{15}$$

$$\mathbf{m}_{n,\text{old}}^{f_n, f_0} = \mathbf{V}_{n,\text{old}}^{f_n, f_0} \left\{\left[\mathbf{V}_{f_n, f_0}^{\mathbf{f}}\right]^{-1} \mathbf{m}_{f_n, f_0}^{\mathbf{f}} - \mathbf{b}_{h_n}\right\}, \tag{16}$$

$$v_{n,\text{old}}^{c_{k,n}} = \left\{\left[v_{n,n}^{\mathbf{c}_k}\right]^{-1} - a_{c_{k,n}}^{h_n}\right\}^{-1}, \tag{17}$$

$$m_{n,\text{old}}^{c_{k,n}} = \left\{m_n^{\mathbf{c}_k} \left[v_{n,n}^{\mathbf{c}_k}\right]^{-1} - b_{h_n}\right\}^{-1}, \tag{18}$$

where $\mathbf{V}^{\mathbf{f}}_{f_n,f_0}$ is the $2 \times 2$ covariance matrix obtained by taking the entries corresponding to $f_n$ and $f_0$ from $\mathbf{V}^{\mathbf{f}}$, and $\mathbf{m}^{\mathbf{f}}_{f_n,f_0}$ is the corresponding 2-dimensional mean vector. Similarly, $v^{\mathbf{c}_k}_{n,n}$ is the variance for $c_{k,n}$ in $\tilde{q}$ and $m^{\mathbf{c}_k}_n$ is the corresponding mean.

To minimize the KL divergence between Eqs. (12) and (13), we match the first and second moments of these two distributions. The moments of Eq. (12) can be obtained from the derivatives of its normalization constant. This normalization constant is given by

$$Z = \int h_n(f_n, f_0, c_{1,n}, \dots, c_{k,n}) \tilde{q}^{\backslash n}(f_n, f_0, c_{1,n}, \dots, c_{k,n}) \, df_n, df_0, dc_{1,n}, \dots, dc_{k,n}$$

$$= \left( \left\{ \prod_{k=1}^{K} \Phi\left[\alpha_n^k\right] \right\} \Phi(\alpha_n) + \left\{ 1 - \prod_{k=1}^{K} \Phi\left[\alpha_n^k\right] \right\} \right), \tag{19}$$

where $\alpha_n^k = m^{c_{k,n}}_{n,\text{old}} / \sqrt{v^{c_{k,n}}_{n,\text{old}}}$ and $\alpha_n = [1, -1]\mathbf{m}^{f_n,f_0}_{n,\text{old}} / \sqrt{[1, -1]\mathbf{V}^{f_n,f_0}_{n,\text{old}}[1, -1]^\top}$. We follow Eqs. (5.12) and (5.13) Minka (2001) to update $a_{h_n}$ and $b_{h_n}$; however, we use the second partial derivatives with respect to $m^{c_{k,n}}_{n,\text{old}}$ rather than first partial derivative with respect to $v^{c_k}_{n,\text{old}}$ for numerical robustness. These derivatives are given by

$$\frac{\partial \log Z}{\partial m^{c_{k,n}}_{n,\text{old}}} = \frac{(Z-1)\phi(\alpha_n^k)}{Z\Phi(\alpha_n^k)\sqrt{v^{c_{k,n}}_{n,\text{old}}}}, \tag{20}$$

$$\frac{\partial^2 \log Z}{\partial [m^{c_{k,n}}_{n,\text{old}}]^2} = -\frac{(Z-1)\phi(\alpha_n^k)}{Z\Phi(\alpha_n^k)} \cdot \frac{\left[\alpha_n^k + \frac{(Z-1)\phi(\alpha_n^k)}{Z\Phi(\alpha_n^k)}\right]}{v^{c^n_t}_{l,\text{old}}}, \tag{21}$$

where $\phi$ and $\Phi$ are the standard Gaussian pdf and cdf, respectively. The update equations for the parameters $a_{h_n}$ and $b_{h_n}$ of the approximate factor $\tilde{h}$ are then

$$a^{\text{new}}_{h_n} = -\left( \left( \frac{\partial^2 \log Z}{\partial [m^{c_{k,n}}_{n,\text{old}}]^2} \right)^{-1} + v^{c_{k,n}}_{n,\text{old}} \right)^{-1},$$

$$b^{\text{new}}_{h_n} = \left\{ m^{c_{k,n}}_{n,\text{old}} - \left[ \frac{\partial^2 \log Z}{\partial [m^{c_{k,n}}_{n,\text{old}}]^2} \right]^{-1} \frac{\partial \log Z}{\partial m^{c_{k,n}}_{n,\text{old}}} \right\} a^{\text{new}}_{h_n} \tag{22}$$

We now perform the analogous operations to update $\mathbf{A}_{h_n}$ and $\mathbf{b}_{h_n}$. We need to compute

$$\frac{\partial \log Z}{\partial \mathbf{m}^{f_n,f_0}_{n,\text{old}}} = \frac{\left\{ \prod_{k=1}^{K} \Phi[\alpha_n^k] \right\} \phi(\alpha_n)}{Z\sqrt{s}}[1, -1],$$

$$\frac{\partial \log Z}{\partial \mathbf{V}^{f_n,f_0}_{n,\text{old}}} = -\frac{1}{2}[1, -1]^\top[1, -1]\frac{\left\{ \prod_{k=1}^{K} \Phi[\alpha_n^k] \right\} \phi(\alpha_n)\alpha_n}{Zs},$$

where

$$s = [-1\,1]\mathbf{V}^{f_n,f_0}_{n,\text{old}}[-1\,1]^\top. \tag{23}$$

We then compute the optimal mean and variance (those that minimize the KL-divergence) of the product in Eq. (13) by using Eqs. (5.12) and (5.13) from Minka (2001):

$$[\mathbf{V}^{\mathbf{f}}_{f_n,f_0}]_{\text{new}} = \mathbf{V}^{f_n,f_0}_{n,\text{old}} - \mathbf{V}^{f_n,f_0}_{n,\text{old}}\left[ \frac{\partial \log Z}{\partial \mathbf{m}^{f_n,f_0}_{n,\text{old}}} \left( \frac{\partial \log Z}{\partial \mathbf{m}^{f_n,f_0}_{n,\text{old}}} \right)^\top - 2\frac{\partial \log Z}{\partial \mathbf{V}^{f_n,f_0}_{n,\text{old}}} \right] \mathbf{V}^{f_n,f_0}_{n,\text{old}},$$

$$[\mathbf{m}^{\mathbf{f}}_{f_n,f_0}]_{\text{new}} = \mathbf{m}^{f_n,f_0}_{n,\text{old}} + \mathbf{V}^{f_n,f_0}_{n,\text{old}}\frac{\partial \log Z}{\partial \mathbf{m}^{f_n,f_0}_{n,\text{old}}}. \tag{24}$$

Next we need to divide the Gaussian with parameters given by Eq. (24) by the Gaussian cavity distribution $\tilde{q}^{\backslash n}(f_n, f_0, c_{1,n}, \ldots, c_{k,n})$. Therefore the new parameters $\mathbf{A}_{h_n}$ and $\mathbf{b}_{h_n}$ of the approximate factor $\tilde{h}$ are obtained using the formula for the ratio of two Gaussians:

$$[\mathbf{A}_{h_n}]^{\text{new}} = \left[\mathbf{V}^{\mathbf{f}}_{f_n, f_0}\right]^{-1}_{\text{new}} - \left[\mathbf{V}^{f_n, f_0}_{n, \text{old}}\right]^{-1} .$$

$$[\mathbf{b}_{h_n}]^{\text{new}} = \left[\mathbf{m}^{\mathbf{f}}_{f_n, f_0}\right]_{\text{new}} \left[\mathbf{V}^{\mathbf{f}}_{f_n, f_0}\right]^{-1}_{\text{new}} - \mathbf{m}^{f_n, f_0}_{n, \text{old}} \left[\mathbf{V}^{f_n, f_0}_{n, \text{old}}\right]^{-1} . \tag{25}$$

## 3.2 EP approximation to $g_k$

We now show how to refine the $\{\tilde{g}_k\}$. We adjust each $\tilde{g}_k$ by minimizing the KL divergence between

$$g_k(c_{k,0})\tilde{q}^{\backslash k}(\mathbf{f}, \mathbf{c}_1, \ldots, \mathbf{c}_K) \tag{26}$$

and

$$\tilde{g}_k(c_{k,0})\tilde{q}^{\backslash k}(\mathbf{f}, \mathbf{c}_1, \ldots, \mathbf{c}_K) , \tag{27}$$

where

$$\tilde{q}^{\backslash k}(\mathbf{f}, \mathbf{c}_1, \ldots, \mathbf{c}_K) = \frac{\tilde{q}(\mathbf{f}, \mathbf{c}_1, \ldots, \mathbf{c}_K)}{\tilde{g}_k(c_{k,0})} .$$

Because $\tilde{q}(\mathbf{f}, \mathbf{c}_1, \ldots, \mathbf{c}_K)$ and $\tilde{g}_k(c_{k,0})$ are both Gaussian, $\tilde{q}^{\backslash k}(\mathbf{f}, \mathbf{c}_1, \ldots, \mathbf{c}_K)$ is also Gaussian. Furthermore, if we marginalize out all variables except those which $\tilde{g}_k$ depends on, namely $c_{k,0}$, then $\tilde{q}^{\backslash k}(\mathbf{f}, \mathbf{c}_1, \ldots, \mathbf{c}_K)$ takes the form $\tilde{q}^{\backslash k}(c_{k,0}) = \mathcal{N}\left(c_{k,0} \mid m^{c_{k,n}}_{k,\text{old}}, v^{c_{k,n}}_{k,\text{old}}\right)$, where, by applying the formula for the ratio of Gaussians, $m^{c_{k,n}}_{k,\text{old}}$ and $v^{c_{k,n}}_{k,\text{old}}$ are given by

$$v^{c_{k,n}}_{k,\text{old}} = \left\{[v^{\mathbf{c}_k}_{n,n}]^{-1} - a_{g_k}\right\}^{-1} ,$$

$$m^{c_{k,n}}_{k,\text{old}} = \left\{m^{\mathbf{c}^k}_n [v^{\mathbf{c}^k}_{n,n}]^{-1} - b_{g_k}\right\}^{-1} . \tag{28}$$

Similarly as in Section 3.1, to minimize the KL divergence between Eqs. (26) and (27), we match the first and second moments of these two distributions. The moments of Eq. (26) can be obtained from the derivatives of its normalization constant. This normalization constant is given by $Z = \Phi(\alpha)$ where $\alpha = m^{c_{k,n}}_{k,\text{old}}/\sqrt{v^{c_{k,n}}_{k,\text{old}}}$. Then, the derivatives are

$$\frac{\partial \log Z}{\partial m^{c_{k,n}}_{k,\text{old}}} = \frac{\phi[\alpha]}{\Phi[\alpha]\sqrt{v^{c_{k,n}}_{k,\text{old}}}}$$

$$\frac{\partial^2 \log Z}{\partial [m^{c_{k,n}}_{k,\text{old}}]^2} = -\frac{\phi[\alpha]}{v^{c_{k,n}}_{k,\text{old}}\Phi[\alpha]} \cdot \left\{\alpha + \frac{\phi[\alpha]}{\Phi[\alpha]}\right\} \tag{29}$$

The update rules for $a_{g_k}$ and $b_{g_k}$ are then

$$[a_{g_k}]^{\text{new}} = -\left(\left[\frac{\partial \log Z}{\partial m^{c_{k,n}}_{k,\text{old}}}\right]^{-1} + v^{c_{k,n}}_{k,\text{old}}\right)^{-1}$$

$$[b_{g_k}]^{\text{new}} = \left\{m^{c_{k,n}}_{k,\text{old}} - \left(\frac{\partial^2 \log Z}{\partial [m^{c_{k,n}}_{k,\text{old}}]^2}\right)^{-1} \frac{\partial \log Z}{\partial m^{c_{k,n}}_{k,\text{old}}}\right\} [a_{g_k}]^{\text{new}} . \tag{30}$$

Once EP has converged we can approximate the NFCPD using Eq. (14) in the main text:

$$p\left(\mathbf{z} \mid \mathcal{D}, \mathbf{x}, \mathbf{x}_\star\right) \approx \int p(z_0 \mid \mathbf{f})p(z_1 \mid \mathbf{c}_1) \cdots p(z_K \mid \mathbf{c}_K) \, \tilde{q}(\mathbf{f}, \mathbf{c}_1, \ldots, \mathbf{c}_K)$$

$$\left(\left\{\prod_{k=1}^{K} \Theta[c_k(\mathbf{x})]\right\} \Theta[f(\mathbf{x}) - f_0] + \left\{1 - \prod_{k=1}^{K} \Theta[c_k(\mathbf{x})]\right\}\right) \, d\mathbf{f} \, d\mathbf{c}_1 \cdots d\mathbf{c}_K, \tag{31}$$

where $\mathbf{z} = (f(\mathbf{x}), c_1(\mathbf{x}), \ldots, c_K(\mathbf{x}))$, the first element in $\mathbf{z}$ is accessed using the index 0 and we have used the assumed independence of the objective and constraints to split $p(\mathbf{z} \mid \mathbf{f}, \mathbf{c}_1, \ldots, \mathbf{c}_K)$ into the product of the factors $p(f(\mathbf{x}) \mid \mathbf{f})$ and $p(c_1(\mathbf{x}) \mid \mathbf{c}_1), \ldots, p(c_K(\mathbf{x}) \mid \mathbf{c}_K)$.

# 4    Performing the integration in Eq. (31)

Because the constraint factor on the right-hand side of Eq. (31) only depends on $f_0$ out of all the integration variables, we can rewrite Eq. (31) as

$$
p\left(\mathbf{z} \mid \mathcal{D}^f, \mathcal{D}^1, \ldots, \mathcal{D}^K, \mathbf{x}, \mathbf{x}_\star^{(m)}\right) \approx \frac{1}{Z} \int \left( \left\{ \prod_{k=1}^{K} \Theta[z_k] \right\} \Theta[z_0 - f_0] + \left\{ 1 - \prod_{k=1}^{K} \Theta[z_k] \right\} \right)
$$
$$
\left( \int p(z_0 \mid \mathbf{f}) \, p(z_1 \mid \mathbf{c}_1) \cdots p(z_K \mid \mathbf{c}_K) \, \tilde{q}(\mathbf{f}, \mathbf{c}_1, \ldots, \mathbf{c}_K) \, df_1 \cdots df_N \, d\mathbf{c}_1 \cdots d\mathbf{c}_K \right) df_0 \, , \tag{32}
$$

where $Z$ is a normalization constant.

We now compute the inner integral in Eq. (32) analytically. This is possible because the marginal posterior predictive distributions on $\mathbf{z}$ are Gaussian and we also have a Gaussian approximation $\tilde{q}(\mathbf{f}, \mathbf{c}_1, \ldots, \mathbf{c}_K)$. We first rewrite the inner integral in Eq. (32) by using the definition of $\tilde{q}$ in Eq. (10):

$$
\int p(z_0 \mid \mathbf{f}) \, p(z_1 \mid \mathbf{c}_1) \cdots p(z_K \mid \mathbf{c}_K) \mathcal{N}(\mathbf{f} \mid \mathbf{m^f}, \mathbf{V^f}) \prod_{k=1}^{K} \mathcal{N}(\mathbf{c}_k \mid \mathbf{m}^{\mathbf{c}_k}, \mathbf{V}^{\mathbf{c}_k}) df_1 \cdots df_N \, d\mathbf{c}_1 \cdots d\mathbf{c}_K \tag{33}
$$

For each $\mathbf{c}_k$, the product of the conditional Gaussian distribution $p(c_k(\mathbf{x}) \mid \mathbf{c}_k)$ with the Gaussian $\mathcal{N}(\mathbf{c}_k \mid \mathbf{m}_k, \mathbf{V}_k)$ is an $(N + 2)$-dimensional multivariate Gaussian distribution. All variables in $\mathbf{c}_k$ are then integrated out leaving only a univariate Gaussian on $c_k(\mathbf{x})$ with mean and variance $m_k'$ and $v_k'$ respectively. For the variables $f(\mathbf{x})$ and $\mathbf{f}$, the product of $p(f(\mathbf{x}) \mid \mathbf{f})$ and $\mathcal{N}(\mathbf{f} \mid \mathbf{m}_0, \mathbf{V}_0)$ yields an $(N + 2)$-dimensional multivariate Gaussian distribution. The variables $f_1, \ldots, f_N$ are integrated out leaving only a bivariate Gaussian on the two-dimensional vector $\mathbf{f}' = (f(\mathbf{x}), f_0)$ with mean vector $\mathbf{m_f'}$ and covariance matrix $\mathbf{V_f'}$. Thus, the result of the inner integral in Eq. (32) is

$$
\int p(z_0 \mid \mathbf{f}) \, p(z_1 \mid \mathbf{c}_1) \cdots p(z_K \mid \mathbf{c}_K) \mathcal{N}(\mathbf{f} \mid \mathbf{m}_0, \mathbf{V}_0) \prod_{k=1}^{K} \mathcal{N}(\mathbf{c}_k \mid \mathbf{m}_k, \mathbf{V}_k) df_1 \cdots df_N \, d\mathbf{c}_1 \cdots d\mathbf{c}_K
$$
$$
= \mathcal{N}\left(\mathbf{f}' \mid \mathbf{m_f'}, \mathbf{V_f'}\right) \prod_{k=1}^{K} \mathcal{N}\left(c_k(\mathbf{x}) \mid m_k', v_k'\right) \, , \tag{34}
$$

where, using Eqs. (3.22) and (3.24) of Rasmussen & Williams (2006) for the means and variances respectively, we have the definitions

$$
[\mathbf{m_f'}]_1 = \mathbf{k}_{\text{final}}^f(\mathbf{x})^\top \left[\mathbf{K}_{\star,\star}^f\right]^{-1} \mathbf{m^f} \, ,
$$
$$
[\mathbf{m_f'}]_2 = \left[\mathbf{m^f}\right]_0 \, ,
$$
$$
[\mathbf{V_f'}]_{1,1} = k_f(\mathbf{x}, \mathbf{x}) - \mathbf{k}_{\text{final}}^f(\mathbf{x})^\top \left\{ \left[\mathbf{K}_{\star,\star}^f\right]^{-1} + \left[\mathbf{K}_{\star,\star}^f\right]^{-1} \mathbf{V^f} \left[\mathbf{K}_{\star,\star}^f\right]^{-1} \right\} \mathbf{k}_{\text{final}}^f(\mathbf{x}) \, ,
$$
$$
[\mathbf{V_f'}]_{2,2} = \left[\mathbf{V^f}\right]_{0,0} \, ,
$$
$$
[\mathbf{V_f'}]_{1,2} = k_f(\mathbf{x}, \mathbf{x}_\star^{(m)}) - \mathbf{k}_{\text{final}}^f(\mathbf{x})^\top \left\{ [\mathbf{K}_{\star,\star}^f]^{-1} + [\mathbf{K}_{\star,\star}^f]^{-1} \mathbf{V^f} [\mathbf{K}_{\star,\star}^f]^{-1} \right\} \mathbf{k}_{\text{final}}^f(\mathbf{x}_\star^{(m)}) \, ,
$$
$$
m_k' = \mathbf{k}_{\text{final}}^k(\mathbf{x})^\top \left[\mathbf{K}_{\star,\star}^k\right]^{-1} \mathbf{m}^{\mathbf{c}_k} \, ,
$$
$$
v_k' = k_k(\mathbf{x}, \mathbf{x}) - \mathbf{k}_{\text{final}}^k(\mathbf{x})^\top \left\{ \left[\mathbf{K}_{\star,\star}^k\right]^{-1} + \left[\mathbf{K}_{\star,\star}^k\right]^{-1} \mathbf{V}^{\mathbf{c}_k} \left[\mathbf{K}_{\star,\star}^k\right]^{-1} \right\} \mathbf{k}_{\text{final}}^k(\mathbf{x}) \, ,
$$

and

- $\mathbf{m^f}$ and $\mathbf{V^f}$ are the posterior mean and posterior covariance matrix of $\mathbf{f}$ given by $\tilde{q}$.

- $\mathbf{k}_{\mathrm{final}}^f(\mathbf{x})$ is the $(N+1)$-dimensional vector with the prior cross-covariances between $f(\mathbf{x})$ and the elements of $\mathbf{f}$ given by $f(\mathbf{x}_\star), f(\mathbf{x}_1), \ldots, f(\mathbf{x}_N)$.

- $\mathbf{K}_{\star,\star}^f$ is an $(N+1) \times (N+1)$ matrix with the prior covariances between elements of $\mathbf{f}$.

- $\mathbf{k}_{\mathrm{final}}^f(\mathbf{x}_\star)$ is the $(N+1)$-dimensional vector with the prior cross-covariances between $f(\mathbf{x}_\star)$ and the elements of $\mathbf{f}$.

- $\mathbf{k}_{\mathrm{final}}^k(\mathbf{x}^k)$ is an $(N+1)$ vector with the cross-covariances between $c_k(\mathbf{x})$ and the elements of $\mathbf{c}_k$ given by $c_k(\mathbf{x}_\star), c_k(\mathbf{x}_1), \ldots, c_k(\mathbf{x}_n)$.

- $\mathbf{K}_{\star,\star}^k$ is an $(N+1) \times (N+1)$ covariance matrix with the prior covariances between the elements of $\mathbf{c}_k$.

We have now computed the inner integral in Eq. (31), leaving us with our next approximation of the NFCPD:

$$
p(f(\mathbf{x}), c_1(\mathbf{x}), \ldots, c_K(\mathbf{x}) \mid \mathcal{D}^f, \mathcal{D}^1, \ldots, \mathcal{D}^K, \mathbf{x}, \mathbf{x}_\star) \approx
$$
$$
\frac{1}{Z} \int \left( \left\{ \prod_{k=1}^{K} \Theta[c_k(\mathbf{x})] \right\} \Theta\left[f(\mathbf{x}) - f_0\right] + \left\{ 1 - \prod_{k=1}^{K} \Theta\left[c_k(\mathbf{x})\right] \right\} \right)
$$
$$
\left\{ \prod_{nk=1}^{K} \mathcal{N}\left(c_k(\mathbf{x}) \mid m_k', v_k'\right) \right\} \mathcal{N}\left(\mathbf{f}' \mid \mathbf{m_f'}, \mathbf{V_f'}\right) \, df_0 \,. \tag{35}
$$

Eq. (35) is the same as Eq. (15) in the main text. Note that the computations performed to obtain $m_k'$, $v_k'$, $\mathbf{m_f'}$, and $\mathbf{V_f'}$ do not depend on $\mathbf{x}$. In the following section we make a final Gaussian approximation to Eq. (35), which must be performed for every $\mathbf{x}$.

# 5   Final Gaussian approximation to the NFCPD, for each $\mathbf{x}$

Because the right-hand side of Eq. (35) is not tractable, we approximate it with a product of Gaussians that have the same marginal means and variances. In particular, we approximate it as

$$
p(\mathbf{z} \mid \mathcal{D}, \mathbf{x}, \mathbf{x}_\star) \approx \mathcal{N}\left(f(\mathbf{x}) \mid \mu_f^{\mathrm{NFCPD}}(\mathbf{x}), v_f^{\mathrm{NFCPD}}(\mathbf{x})\right) \prod_{k=1}^{K} \mathcal{N}\left(c_k(\mathbf{x}) \mid \mu_k^{\mathrm{NFCPD}}(\mathbf{x}), v_k^{\mathrm{NFCPD}}(\mathbf{x})\right) \,,
$$

where $\mu_{\mathrm{NFCPD}}^f(\mathbf{x})$, $v_{\mathrm{NFCPD}}^f(\mathbf{x})$ and $\mu_{\mathrm{NFCPD}}^k(\mathbf{x})$ and $v_{\mathrm{NFCPD}}^k(\mathbf{x})$ are the marginal means and marginal variances of $f(\mathbf{x})$ and $c_k(\mathbf{x})$ according to the right-hand-side of Eq. (35). Using Eqs. (5.12) and (5.13) in Minka (2001) we can compute these means and variances in terms of the derivatives of the normalization constant $Z$ in Eq. (35), which is given by

$$
Z = \left\{ \prod_{k=1}^{K} \Phi[\alpha_k] \right\} \Phi[\alpha] + \left\{ 1 - \prod_{k=1}^{K} \Phi[\alpha_k] \right\} \tag{36}
$$

where

$$
s = [\mathbf{V_f'}]_{1,1} + [\mathbf{V_f'}]_{2,2} - 2[\mathbf{V_f'}]_{1,2}
$$
$$
\alpha = \frac{[1, -1]\mathbf{m_f'}}{\sqrt{s}}
$$
$$
\alpha_k = \frac{m_k'}{\sqrt{v_k'}} \,.
$$

Doing so yields

$$v_f^{\text{NFCPD}}(\mathbf{x}) = [\mathbf{V'_f}]_{1,1} - \frac{\beta}{s}\,(\beta + \alpha)\left([\mathbf{V'_f}]_{1,1} - [\mathbf{V'_f}]_{1,2}\right)^2$$

$$\mu_f^{\text{NFCPD}}(\mathbf{x}) = [\mathbf{m'_f}]_1 + \left([\mathbf{V'_f}]_{1,1} - [\mathbf{V'_f}]_{1,2}\right)\frac{\beta}{\sqrt{s}}$$

$$v_k^{\text{NFCPD}}(\mathbf{x}) = \left\{[v'_k]^{-1} + \tilde{a}\right\}^{-1}$$

$$\mu_k^{\text{NFCPD}}(\mathbf{x}) = v_{N,\text{const}}^k(\mathbf{x})\left\{[m'_k]^{-1}[v'_k]^{-1} + \tilde{b}\right\}, \tag{37}$$

where

$$\beta = \frac{\left\{\prod_{k=1}^{K}\Phi[\alpha_k]\right\}\phi(\alpha)}{Z}$$

$$\tilde{a} = -\left\{\frac{\partial^2\log Z}{\partial\,[m'_k]^2} + v'_k\right\}^{-1}$$

$$\tilde{b} = \tilde{a}\left\{m'_k + \frac{\sqrt{v'_k}}{\alpha_k + \beta_k}\right\}$$

$$\beta_k = \frac{\phi(\alpha_n)}{Z\Phi(\alpha_n)}(Z-1),$$

$$\frac{\partial^2\log Z}{\partial\,[m'_k]^2} = -\frac{\beta_k\{\alpha_k + \beta_k\}}{v'_k}.$$

## 5.1 Initialization and convergence of EP

Initially EP sets the parameters of all the approximate factors to be zero. We use at the convergence criterion that the absolute change in all parameters should be below $10^{-4}$.

## 5.2 EP with damping

To improve convergence we use damping (Minka & Lafferty, 2002). If $\tilde{h}_n^{\text{new}}$ is the minimizer of the KL-divergence, damping entails using instead $\tilde{h}_n^{\text{damped}}$ as the new factor, as defined below:

$$\tilde{h}_n^{\text{damped}} = [\tilde{h}_n^{\text{new}}]^\epsilon + \tilde{h}_n^{1-\epsilon}, \tag{38}$$

where $\tilde{h}_n$ is the factor at the previous iteration. We do the same for $\tilde{g}_k$. The parameter $\epsilon$ controls the amount of damping, with $\epsilon = 1$ corresponding to no damping. We initialize $\epsilon$ to 1 and multiply it by a factor of 0.99 at each iteration. Furthermore, the factors $\tilde{h}_n$ and $\tilde{g}_k$ are updated in parallel (i.e. without updating $\tilde{q}$ in between) in order to speed up convergence (Gerven et al., 2009). During the execution of EP, some covariance matrices may become non positive definite due to an excessively large step size (i.e. large $\epsilon$). If this issue is encountered during an EP iteration, the damping parameter is reduced by half and the iteration is repeated. The EP algorithm is terminated when the absolute change in all the parameters in $\tilde{q}$ is less than $10^{-4}$.



\end{document}